\documentclass[a4paper,fleqn]{cas-dc}

\usepackage[numbers]{natbib}

\usepackage{url}
\usepackage{verbatim}
\usepackage{times}
\usepackage{epsfig}
\usepackage{graphicx}
\usepackage{amsmath,amsfonts,amssymb,amsthm,version}
\usepackage{algorithmic}
\usepackage{algorithm}
\usepackage{array}
\usepackage{tikz-cd}

\theoremstyle{}

\newtheorem{definition}{Definition}

\theoremstyle{remark}

\begin{document}
\let\WriteBookmarks\relax
\def\floatpagepagefraction{1}
\def\textpagefraction{.001}

\shorttitle{}    

\shortauthors{}  

\title [mode = title]{Topology Structure Optimization of Reservoirs Using GLMY Homology}  



\author[1]{Yu Chen}
\ead{chenyu.math@zju.edu.cn}

\author[2]{Shengwei Wang}
\ead{wsw_817@zju.edu.cn}

\author[1]{Hongwei Lin}
\cormark[1]
\ead{hwlin@zju.edu.cn}

\affiliation[1]{organization={School of Mathematical Sciences, Zhejiang University},
	city={Hangzhou},
	postcode={310058},
	country={China}}
    
\affiliation[2]{organization={Polytechnic Institute of Zhejiang University},
	city={Hangzhou},
	postcode={310015},
	country={China}}
    
\cortext[1]{Corresponding author}



\begin{abstract} 
	Reservoirs are efficient networks for time-series processing. 
	It is well known that the network structure is one of the determinants of their performance. 
	However, the topological structure of reservoirs, 
		as well as their performance,
		is hard to analyze
		due to the lack of suitable mathematical tools. 
	In this paper, we study the topological structure of reservoirs using 
		persistent GLMY homology theory 
		and develop a method to improve their performance.
	Specifically, we find that reservoir performance is correlated with the
		one-dimensional GLMY homology groups.
	Then, we develop a reservoir structure optimization method by modifying 
		the minimal representative cycles of one-dimensional GLMY homology groups.
	Finally, through experiments,
		we validate that the performance of reservoirs is jointly influenced by the reservoir
		structure and the periodicity of the dataset. 
	
\end{abstract}


\let\printorcid\relax
\maketitle

\section{Introduction}
Neural networks, inspired by biological brains, have become a cornerstone in artificial intelligence and machine learning, enabling solutions to complex tasks in image recognition, natural language processing, and predictive analytics \cite{lecun2015deep,goodfellow2016deep}. Their structure—including layer count, neuron connectivity, and activation function type—directly determines functional capabilities, with early studies establishing foundational concepts like weight roles and nonlinearity from activation functions \cite{hornik1989multilayer}. Moreover, a recent study has proposed weight-agnostic neural networks \cite{gaier2019weight}, which can perform learning tasks without weight training, relying only on structural optimization. This shows that the structure of neural networks plays an essential role. However, as real-world problem complexity grows, a deeper understanding of task-optimized neural network structure design is needed.

The continuous improvement and optimization of neural network structures hold profound and far-reaching significance in advancing the practical application and theoretical development of artificial intelligence. From a performance perspective, a well-optimized structure can significantly enhance the network's ability to extract features and learn patterns, enabling it to achieve higher accuracy and robustness in handling complex tasks. In terms of resource efficiency, optimizing the number of layers, neuron density, and connectivity patterns can effectively reduce the model's parameter scale and computational complexity. Additionally, structural optimization plays a crucial role in enhancing the adaptability and interpretability of neural networks. By adjusting the structure to match the characteristics of specific tasks, the network's generalization ability across different domains can be improved. Meanwhile, rational structural design can reduce the ``black-box'' nature of neural networks—for instance, through modular structural design, researchers can more clearly trace the flow of information and the contribution of each module to the final output, laying the foundation for interpreting the internal working mechanism of the network \cite{montavon2018methods}. Moreover, recent studies indicate that the topology of a network's structure significantly impacts its performance, suggesting that understanding and designing network architectures from a topological perspective is a crucial and effective approach \cite{zhang2024comprehensive}.

Reservoir Computing (RC), as a distinctive paradigm within the machine learning landscape, has rapidly gained traction and widespread adoption owing to its inherent advantages that address key challenges in traditional neural network deployment. Specifically, RC exhibits a compact model architecture, which minimizes the demand for storage resources; enables rapid training processes by focusing optimization only on the readout layer rather than the entire network, significantly reducing computational overhead; delivers high prediction accuracy across various time-series and nonlinear tasks; and maintains excellent generalization performance when faced with unseen or noisy data—these attributes collectively make RC a preferred choice for applications ranging from real-time signal processing to complex system prediction \cite{lukovsevivcius2009reservoir,jaeger2001echo}. Nevertheless, a critical limitation of RC lies in its ``black-box'' nature, primarily stemming from the lack of clear interpretability regarding the internal dynamics of the reservoir. The reservoir, typically composed of a large number of randomly connected neurons, operates through complex nonlinear interactions, making it difficult to establish a direct link between its structural characteristics (e.g., connectivity density, neuron type, and topological arrangement) and its prediction performance. Consequently, investigating how the internal structure of the reservoir influences its prediction ability has become an increasingly attractive and essential research direction, as resolving this issue could not only unlock the potential to further optimize RC performance but also enhance its reliability and applicability in critical domains where interpretability is essential \cite{pathak2018model}.

In practice, a reservoir is usually regarded as a digraph. 
In previous studies, the lack of mathematical tools to study the topology of digraphs made it difficult to topologically optimize the structure of the reservoir from an algebraic topological perspective. Now with the rapid development of \textit{GLMY homology} theory, there are new effective tools for topology optimization of digraphs.
The GLMY homology, also known as path homology and initially proposed by \cite{grigor2012homologies}, represents a significant development in computational topology. As complex systems—ranging from social networks and neural connections to transportation and biochemical pathways—are increasingly modeled using directed graphs (digraphs), traditional homology theories, which are designed for undirected topological spaces, prove inadequate. They fail to capture crucial directional information, such as information flow, influence propagation, or causal relationships. GLMY homology overcomes this fundamental limitation by constructing a chain complex specifically based on directed paths within digraphs. Its distinct sensitivity to edge directions allows it to capture the flow and hierarchical organization of these paths, revealing hidden structural features and topological properties that are otherwise invisible to standard methods. In real-world digraphs, edge orientation often conveys critical information, such as influence direction in a social network or material flow in a manufacturing process. The directional awareness of GLMY homology provides a more accurate and nuanced representation of the digraph's underlying topology. Persistent GLMY homology \cite{chowdhury2018persistent}, a further development of GLMY homology, enhances its analytical capabilities by tracking how directed path structures evolve as parameters, such as edge weights, change. Through persistent GLMY homology, researchers can effectively capture structural features encoded in both edge directions and weights, making it a powerful tool for analyzing dynamic and weighted complex systems. This framework provides a new lens through which to analyze the structural properties of complex networks, particularly those where directionality is a defining feature.

Therefore, in this study, since a reservoir can be represented as a digraph, we present a novel method for reservoir optimization using GLMY homology. Firstly, we show that adding \textit{rings} \footnote{While some literature uses the term ``cycles'', we adopt ``rings'' to avoid conceptual conflicts with homology theory.} in the reservoir digraph can improve the orthogonality of the reservoir adjacency matrix, hence enhancing the reservoir's predictive capacity. 
Mathematically, a directed ring structure with uniform weights corresponds to a (scaled) permutation adjacent matrix. Permutation matrices are a fundamental subclass of orthogonal matrices. As highlighted in \cite{gallicchio2019reservoir}, structured reservoirs based on permutation matrices provide a particularly advantageous architectural setting by ensuring stable signal propagation and mitigating the vanishing gradient problem through their near-orthogonal nature.
Moreover, Rodan and Tino \cite{rodan2010minimum} introduced the Simple Cycle Reservoir (SCR), proving that a deterministic ring topology is sufficient to capture complex temporal features and can reach near-optimal memory capacity. And theoretical studies in \cite{stockdill2016restricted} also confirm that rings are essential for replicating the memory-leaking and state-holding properties required for temporal processing. By increasing the number of rings, one can effectively increase the number of paths for information to recirculate, which provably maintains the ``fading memory'' property over longer time horizons compared to purely random connections.
Hence, these studies serve as the fundamental motivation for our approach. Our strategy focuses on strategically modifying the edge directions within the digraph to deliberately increase the number of rings, thereby optimizing the reservoir's performance. Moreover, since these rings are a special type of representative cycle of one-dimensional GLMY homology, we can achieve our target by considering a subset of representative cycles of 1D GLMY homology, i.e., the subset consists of all rings.
In other words, our optimization strategy, which focuses on increasing the number of rings, is theoretically motivated by the advantages of cyclic topologies. According to the SCR theory \cite{rodan2010minimum}, rings provide the necessary feedback loops for stable fading memory. By leveraging GLMY homology, we provide a systematic way to enhance these critical cyclic structures in an initially random reservoir, effectively moving its topology toward a more organized, high-memory-capacity state.
We also validate through experiments that the prediction ability of a reservoir is related to both its structure and the topological features of the dataset.

In summary, the main contributions of our study are as follows:
\begin{itemize}
    \item Adding rings inside the reservoir digraph can improve the orthogonality of the reservoir adjacency matrix, which is equivalent to increasing a certain type of representative cycle of GLMY homology.
    \item We optimize the structure of reservoirs by expanding a certain subset of representative cycles (rings) of 1D GLMY homology.
    \item It is validated by experiments that the performance of reservoirs is related to both their structure and the characteristics of the dataset. 
\end{itemize}

The remainder of this paper is organized as follows: We provide an overview of reservoir computing, reservoir optimization and GLMY homology in Section \ref{sec:RW}. Then, we present the key tools and the proposed method in Section \ref{sec:preliminary} and \ref{sec:methods}. In Section \ref{sec:exp}, we show our experimental results and provide some discussions. Section \ref{sec:conclusion} presents the conclusions of this study.

\begin{figure*}[!t]
    \centering
    \includegraphics[width=0.95\linewidth]{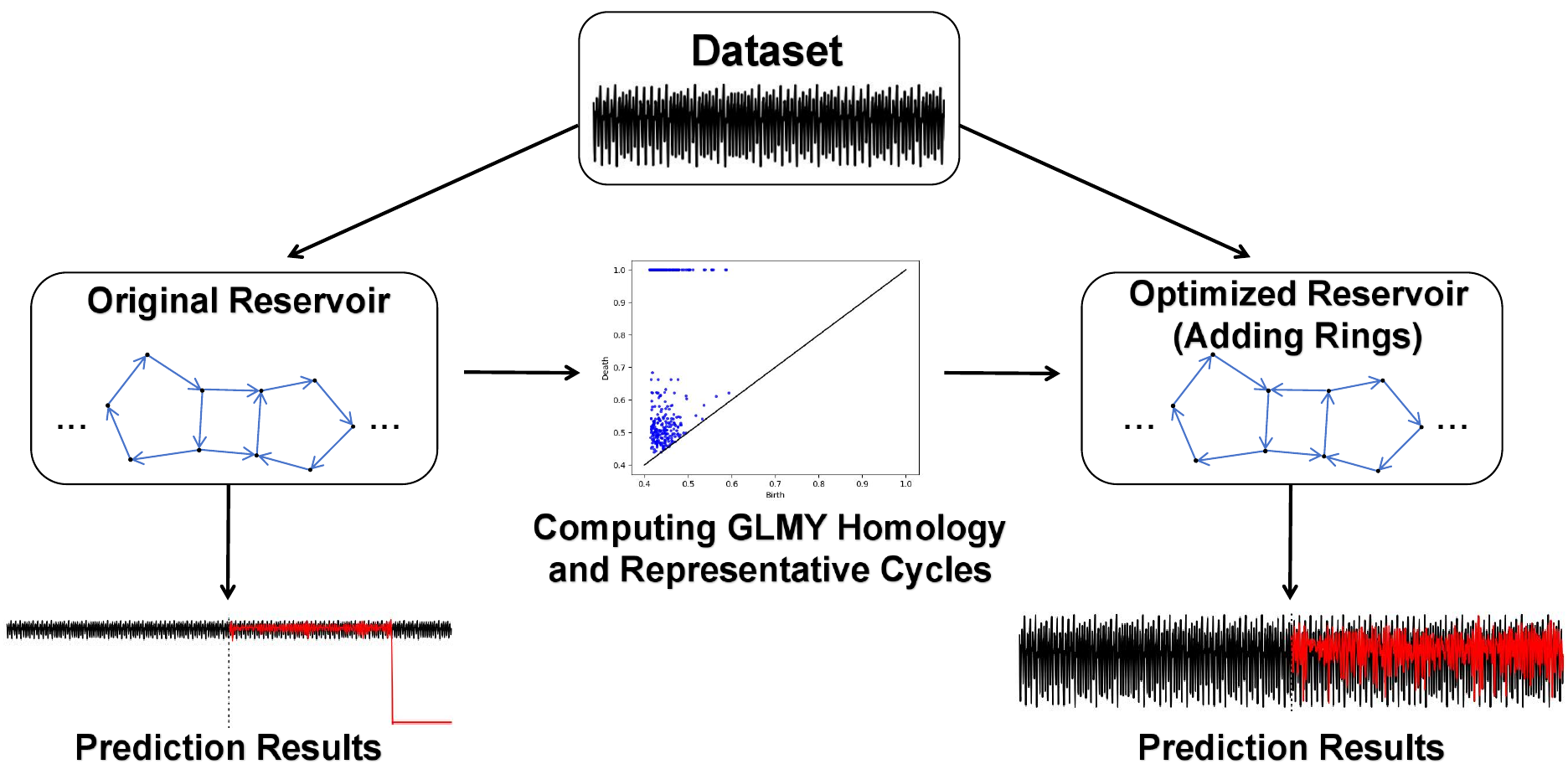}
    \caption{Illustration of the flowchart of our method and experimental procedures.}
    \label{fig:flowchart}
\end{figure*}

\section{Related Work}\label{sec:RW}
In this section, we provide some related work about optimizing reservoir structure, the GLMY homology, and some of its applications.

\subsection{Reservoir computing and reservoir optimization}
Reservoir computing (RC) has emerged as a powerful and widely adopted machine learning paradigm over the past few decades, gaining considerable attention for its effectiveness in handling complex time-dependent data. Its application has extended across various domains, including time series prediction, nonlinear system identification~\cite{yao2019prediction}, and speech recognition~\cite{skowronski2007noise}. The core strength of RC lies in its architectural simplicity: a fixed, randomly generated recurrent neural network (the reservoir) projects the input into a high-dimensional feature space, while only a simple linear readout layer is trained. This separation of the nonlinear dynamics of the reservoir from the linear readout training process is a key reason for its exceptional computational efficiency and rapid training speed. This efficiency has led to remarkable results, such as those demonstrated by Jaeger et al. \cite{jaeger2004harnessing}, who applied RC to chaotic time series prediction and achieved computational accuracy two orders of magnitude higher than traditional approaches. Beyond its core architecture, variations have further expanded its capabilities. For instance, Shougat et al. \cite{shougat2021hopf} employed Hopf oscillators as an intermediate reservoir, successfully demonstrating robustness to noise and preservation of predictive accuracy even with reduced training costs. In the financial sector, Wang et al. \cite{wang2021stock} utilized an RC network for forecasting stock market prices, where comparative experiments against several classical models—including long short-term memory (LSTM) networks—showed the RC model consistently achieving lower mean-squared error (MSE) and mean absolute error (MAE), alongside higher $R^2$ scores. This work also provided insights into how distinct reservoir initialization strategies influence predictive accuracy, further highlighting the critical role of network dynamics in financial forecasting.

A reservoir is usually represented as a digraph (called the reservoir digraph). In most implementations, reservoir parameters and topological configurations are initialized stochastically. However, such stochastic initialization frequently produces unstable dynamical behavior, necessitating extensive empirical trial‑and‑error to obtain a well‑conditioned network. Therefore, beyond investigating the deployment of reservoirs in prediction tasks, the design of high‑quality reservoir architectures has emerged as both a major challenge and a focal point of contemporary research. 
Gallicchio et al. \cite{gallicchio2019reservoir} assessed the impact of constrained reservoir topologies—namely permutation, ring, and chain structures—within Deep Echo State Networks (DeepESNs) on time series forecasting performance. Through a systematic comparison against conventional sparsely random reservoirs and shallow ESNs, demonstrated that embedding structurally sparse topologies in deep reservoir computing constitutes a straightforward yet effective strategy for enhancing temporal modeling accuracy. 
Cui et al. \cite{cui2012architecture} investigated the interplay between reservoir topology and predictive performance by constructing reservoirs with small‑world, scale‑free, and hybrid small‑world/scale‑free structures. Experimental results demonstrate that manually designed reservoir networks deliver superior predictive accuracy and possess a broader spectral radius, while their short‑term memory capacity remains comparable to that of traditional models.
Moreover, Roeschies et al. \cite{roeschies2010structure} proposed an evolutionary algorithm that optimizes reservoir‑network architectures for specific problem classes, resulting in substantial gains in predictive performance. 
Sun et al. \cite{sun2012modeling} introduced a deterministic echo state network architecture featuring a loop reservoir with adjacent feedback connections. This simplified design builds upon the basic loop reservoir structure by incorporating regular adjacent feedback, requiring optimization of only a single parameter. The approach significantly reduces the complexity of ESN implementation while maintaining computational capabilities.
These results, together with previous studies, confirm that the structure of a reservoir is pivotal in determining its functionality and overall effectiveness.

In a broader optimization context, surrogate-based and response surface methodologies (RSM) have been extensively studied in engineering design and system optimization. These approaches typically construct an explicit or implicit approximation of a performance landscape over a continuous parameter space, and subsequently perform optimization on the learned surrogate model.
Representative examples include \cite{ahmed2024experimental,ahmed2025experimental}, where response surfaces are employed to guide design optimization under computational or experimental constraints. While effective for continuous and parameterized systems, such approaches fundamentally differ from our concern: the reservoir structure considered in this work is discrete, graph-based, and non-differentiable.
We will focus on topology-level structural modifications guided by algebraic-topological invariants, providing a complementary optimization paradigm that is particularly suitable for discrete dynamical networks.

\subsection{GLMY homology and its applications}
GLMY homology represents a significant extension of classical homology theory, specifically developed to analyze directed or asymmetric interactions in digraphs \cite{grigor2012homologies} and hypergraphs \cite{grigor2019homology}. This framework has spawned several theoretical developments, such as homotopy theory for digraphs \cite{grigor2014homotopy}, discrete Morse theory on digraphs \cite{lin2021discrete}, and path homology theory for multigraphs and quivers \cite{grigor2018path}.
Notable computational advances include the persistent path homology for networks proposed by Chowdhury and Mémoli \cite{chowdhury2018persistent}, which offers a novel approach to digraph analysis in topological data analysis, and an efficient algorithm for computing one-dimensional persistent path homology of digraph filtrations by Dey et al. \cite{dey2022efficient}. 
In contrast to traditional simplicial complexes, GLMY homology offers a unique set of dimensional fingerprints. These fingerprints are instrumental in unearthing the intrinsic structural information hidden within networks, providing a more detailed view of network architecture.
This theoretical framework has demonstrated its practical utility across a diverse range of applications, such as modeling intricate biological systems \cite{feng2024hypernetwork}, analyzing material properties and structures \cite{liu2023neighborhood}, analyzing discrete planar vector fields \cite{chen2025analyzing}, and understanding complex networks \cite{chowdhury2019path,chowdhury2020path}.
Persistent GLMY homology represents a further advancement in this area. By tracking how topological features change and evolve across various filtration processes, it allows us to identify and analyze topological patterns in networks \cite{chowdhury2018persistent}. This capability significantly enhances our understanding of network behavior and evolution.

\section{Preliminary}\label{sec:preliminary}
In this section, basic concepts of digraphs and persistent homology are introduced.

A \textit{digraph} is an ordered pair $G = (V,E)$, where $V$ denotes the vertex set and $E \subseteq V \times V$ represents the directed edges. A digraph without loops or multiple edges is called a \textit{simple digraph}.
The GLMY homology (or path homology) provides a topological framework for analyzing simple digraphs \cite{grigor2012homologies}. For a finite nonempty vertex set $V$, an \textit{elementary $n$-path} is defined as a sequence $e_{i_0 \ldots i_n}$ of $n+1$ vertices, where $n \geq 0$. By convention, the set of elementary $(-1)$-paths is empty. The boundary operator acts on elementary $n$-paths as:
$$
\partial e_{i_0 \ldots i_n} = \sum_{k=0}^{n} (-1)^k e_{i_0 \ldots \hat{i}_k \ldots i_n},
$$
where $\hat{i}_k$ indicates the omission of vertex $i_k$ from the sequence.
An elementary path $e_{i_0 \ldots i_n}$ on a set $V$ is \textit{regular} if $i_{k-1} \neq i_k, \forall k=1, \ldots, n$. Otherwise, it is \textit{non-regular}. Thus, using the vertex set of a simple digraph, we can obtain the set of elementary $n$-paths of $G$, where all elements are regular paths. In the following discussion, unless otherwise noted, simple digraphs are considered.

Let $G = (V,E)$ be a simple digraph. For any integer $n \geq 0$, we define the $\mathbb{K}$-linear space of \textit{allowed $n$-paths} as:
$$
\mathcal{A}_n(G) = \operatorname{span}\left\{e_{i_0 \ldots i_n} : i_k i_{k+1} \in E \text{ for } k = 0,\ldots,n-1\right\},
$$
where $e_{i_0 \ldots i_n}$ denotes an elementary $n$-path. For $n \geq -1$, the subspace of \textit{$\partial$-invariant $n$-paths} is defined as:
$$
\Omega_n(G) = \left\{v \in \mathcal{A}_n : \partial v \in \mathcal{A}_{n-1}\right\},
$$
with the boundary cases $\Omega_{-1} \cong \mathbb{K}$ and $\Omega_{-2} = \{0\}$. This gives rise to the chain complex:
$$
\cdots \xrightarrow{\partial} \Omega_3 \xrightarrow{\partial} \Omega_2 \xrightarrow{\partial} \Omega_1 \xrightarrow{\partial} \Omega_0 \xrightarrow{\partial} \mathbb{K} \xrightarrow{\partial} 0.
$$
The \textit{$n$-dimensional GLMY homology group} of $G$ is then:
$$
H_n(G) = \operatorname{Ker}(\partial|_{\Omega_n}) / \operatorname{Im}(\partial|_{\Omega_{n+1}}).
$$
Elements of $Z_n := \operatorname{Ker}(\partial|_{\Omega_n})$ are called \textit{$n$-cycles}, while elements of $B_n := \operatorname{Im}(\partial|_{\Omega_{n+1}})$ are \textit{$n$-boundaries}. A \textit{representative} of a homology generator is any $n$-cycle that is not an $n$-boundary, though such representatives are generally non-unique.

\begin{figure*}[!t]
    \centering
    \includegraphics[width=0.85\linewidth]{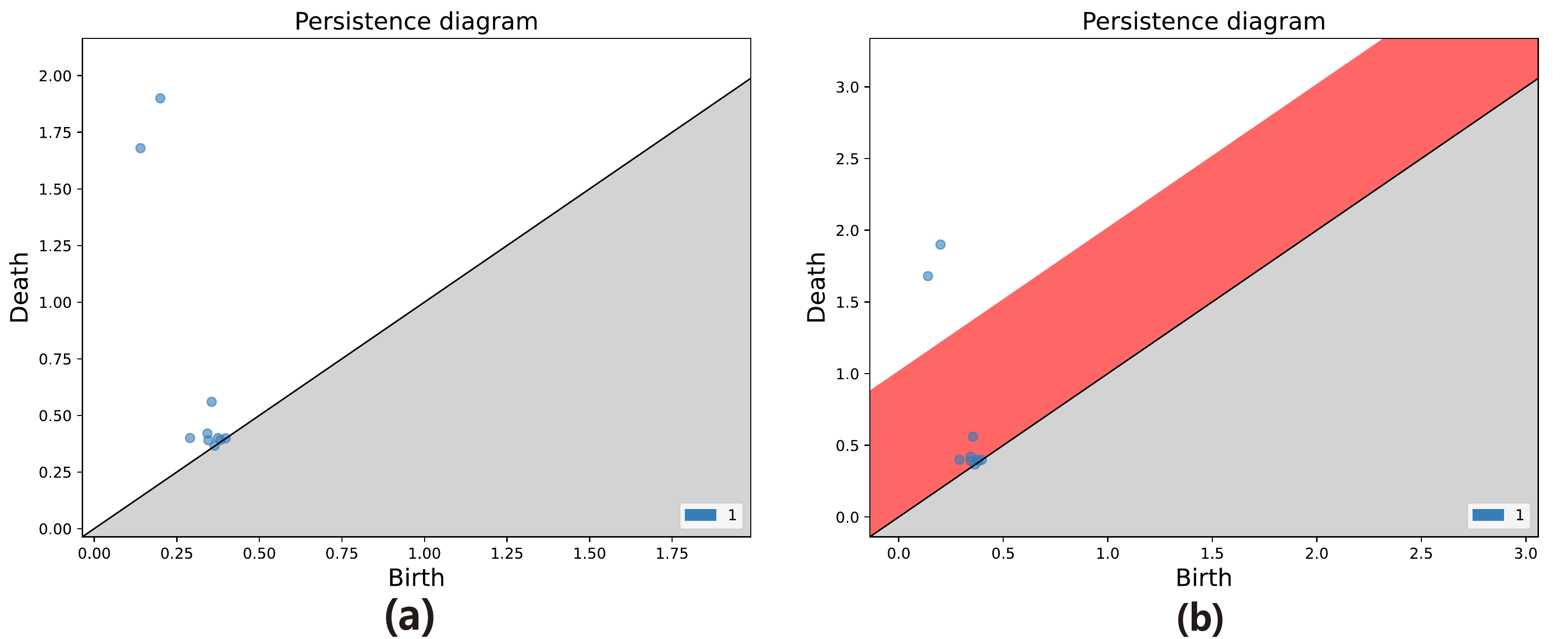}
    \caption{(a) Example of a one-dimensional persistence diagram. (b) An illustration of the confidence set (red area) of (a).}
    \label{fig:PD}
\end{figure*}

Persistent homology is a tool from computational topology that captures the topological features of space at various scales \cite{edelsbrunner2008persistent}. It provides a way to quantify the persistence of topological features, capturing the birth and death of topological features in the considered space (such as point clouds, graphs, and digraphs) by constructing a filtration.
The \textit{persistence diagram} (PD) serves as a powerful visualization tool for persistent homology, where each point $(b_i, d_i)$ corresponds to a \textit{persistence pair} representing the birth ($b_i$) and death ($d_i$) times of a homology generator (Fig. \ref{fig:PD}). Notice that all points in a PD are located above or on the diagonal. And the value $|d_i-b_i|$ is called the \textit{persistence} of the point $(b_i, d_i)$, representing the significance of the corresponding topological features. The greater the persistence of a point in the persistence diagram, the more significant the corresponding topological feature is. 

A statistical approach for identifying significant points in a persistence diagram is the \textit{confidence set} \cite{fasy2014confidence}, which categorizes these points into two classes based on their persistence. As shown in Fig. \ref{fig:PD} (b), the confidence set draws two areas on the persistence diagram: the red area and the white area, which contain points with less persistence and greater persistence, respectively. Points in the class that show greater persistence are called \textit{significant points}, while the other points are called \textit{noise points}.

The tool of persistence diagram is first employed to illustrate the persistent homology computed from some point clouds, and this framework can also naturally extend to persistent GLMY homology, where we denote the one-dimensional persistence diagram of $H_1$ as $PD_1$.

\section{Methodology}\label{sec:methods}
In this section, we first introduce some basic concepts and methods of reservoir computing, as well as some theoretical analysis. In particular, we consider optimization by improving the orthogonality of the reservoir digraph matrix, which enhances its memory capacity. To achieve this goal, the GLMY theory will be used. Subsequently, we further introduce the persistent GLMY homology of a digraph and the proposed approach for optimizing the reservoirs.

\subsection{Reservoir computing and its theoretical analysis}\label{sec:4.1}
\begin{figure*}[!t]
    \centering
    \includegraphics[width=0.6\linewidth]{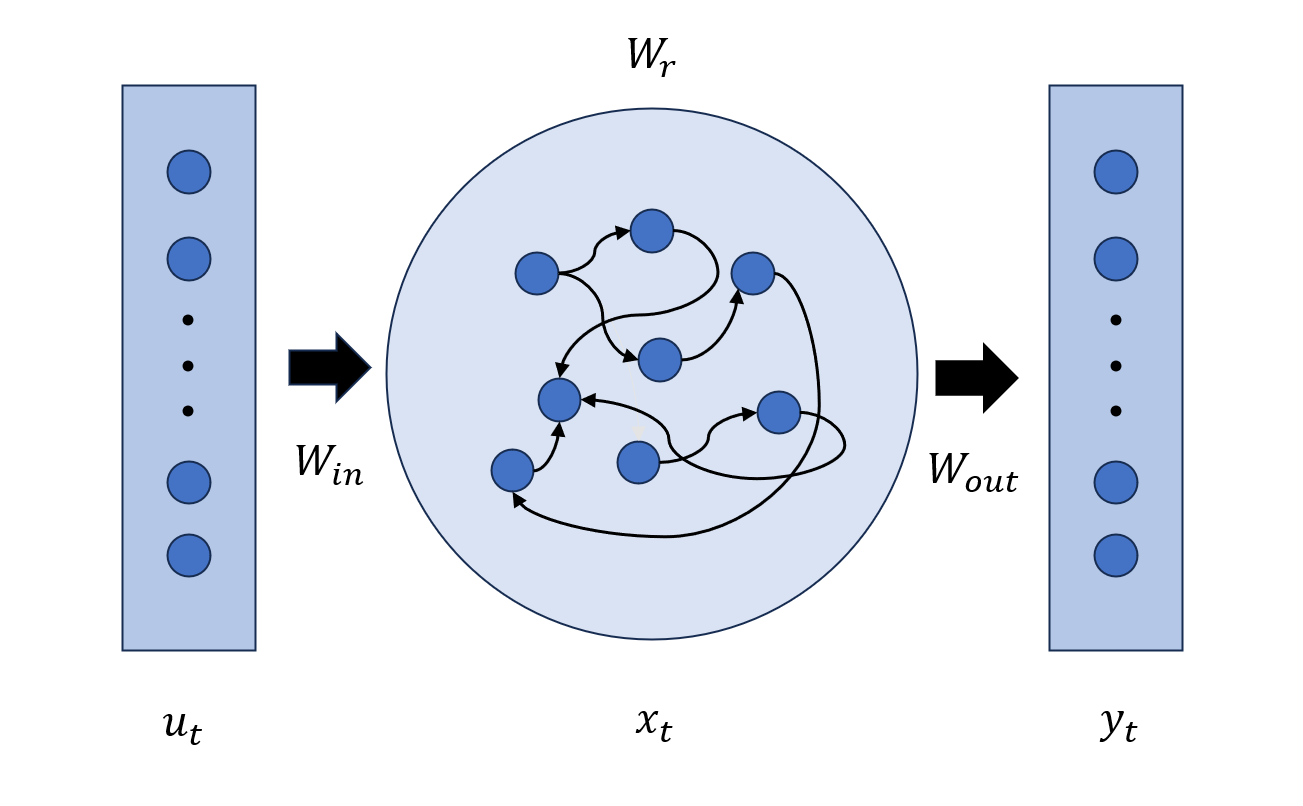}
    \caption{Structure of Reservoir Computing Model}
    \label{fig:Struct of reservoir}
\end{figure*}

Training the weight matrices in deep learning models is typically computationally expensive. In contrast to conventional neural networks, where the weights of all layers are trained, the reservoir computing (RC) architecture only requires the training of the readout layer's weights. This characteristic significantly reduces the computational complexity, offering a computational efficiency advantage in certain applications.
The general structure of the reservoir computing model is illustrated in Figure \ref{fig:Struct of reservoir}. It primarily consists of three components: the input layer, the reservoir, and the output layer. The input layer is responsible for encoding the incoming data and feeding it into the reservoir. The reservoir consists of a large, fixed, and recurrent network of neurons that project the input data into a high-dimensional dynamic space. The output layer, which is typically trained, maps the high-dimensional representations from the reservoir to the desired output. Generally, the input and output layers are fully connected layers, while the reservoir is represented by a weight matrix \( W_{\text{r}} \in \mathbb{R}^{N \times N} \), where \( N \) denotes the size of the reservoir. The weight matrix defines the connections between the neurons within the reservoir, determining the structure and dynamics of the recurrent network. Similarly, the input layer \( L_{\text{in}} \) and the output layer \( L_{\text{output}} \) are also described by their respective weight matrices. Specifically, the input layer is represented by the weight matrix \( W_{\text{in}} \in \mathbb{R}^{N \times L} \), where \( L \) is the dimensionality of the input data. The output layer is represented by the weight matrix \( W_{\text{out}} \in \mathbb{R}^{M \times N} \), where \( M \) is the number of output neurons. These weight matrices govern the flow of information from the input to the reservoir and from the reservoir to the output layer, ensuring the appropriate transformations and mappings of data within the network. Typically, both \( W_{\text{in}} \) and \( W_{\text{r}} \) are randomly initialized and remain fixed throughout operation, whereas \( W_{\text{out}} \) is optimized via gradient descent or ordinary least‑squares methods to obtain the optimal parameters. This strategy substantially reduces the training time and computational resources.

At each discrete time step \(t+1\), the new reservoir state is jointly determined by the external input arriving at that moment and the internal state carried over from the previous step. Concretely, the update rule is often written as
\begin{equation}
\mathbf{x}_{t+1}
= (1-\alpha)\mathbf{x}_{t}
+ \alpha\phi\bigl(
  W_{\text{in}}\mathbf{u}_{t+1}
  + W_{\text{r}}\mathbf{x}_{t}
  + \mathbf{b}
\bigr),
\label{eq:RC1}
\end{equation}
where \(\mathbf{u}_{t+1}\in\mathbb{R}^{L}\) is the input vector, \(\mathbf{x}_{t}\in\mathbb{R}^{N}\) is the previous reservoir state,  
\(W_{\text{in}}\in\mathbb{R}^{N\times L}\) and \(W_{\text{r}}\in\mathbb{R}^{N\times N}\) are the fixed input and reservoir weight matrices, \(\mathbf{b}\) is an optional bias, and \(\phi(\cdot)\) is a point‑wise non‑linear activation (e.g., \(\tanh\)).  
The leak rate \(\alpha\) controls the trade‑off between retaining historical information and incorporating new information.

More precisely, the activation of each neuron inside the reservoir at time step \(t+1\) depends on (i) the subset of input‑layer neurons projecting to it and (ii) the neighbouring reservoir neurons to which it is connected.  
Denoting the \(i\)-th reservoir neuron by \(x_{i,t}\), we have
\begin{equation}
x_{i,t+1} =
(1-\alpha)x_{i,t}
+
\alpha\phi\Bigl(
\sum_{j=1}^{L}  W_{\text{in},ij}u_{j,t+1}
+
\sum_{k=1}^{N}  W_{\text{r},ik}x_{k,t}
+ b_{i}
\Bigr),
\label{eq:RC2}
\end{equation}
where \(u_{j,t+1}\) is the \(j\)-th component of the input vector, \(W_{\text{in},\,ij}\) and \(W_{\text{r},\,ik}\) are the fixed weights onto neuron \(i\), and \(b_{i}\) is an optional bias term. 

Inspection of the update rules in Eqs.~\eqref{eq:RC1}--\eqref{eq:RC2} reveals that, much like a conventional recurrent neural network (RNN)\cite{rumelhart1986learning}, the reservoir maintains a compact representation of both historical and instantaneous information within its state \(\mathbf{x}_{t}\). This internal memory allows the model to integrate long‑range temporal dependencies when predicting future outcomes. The output at time \(t\) is produced by the current reservoir state:
$$
\mathbf{y}_{t}
= W_{\mathrm{out}}\,\mathbf{x}_{t}
+ \mathbf{c},
\label{eq:RC3}
$$
where \(\mathbf{y}_{t}\in\mathbb{R}^{M}\) denotes the output vector,  
\(W_{\mathrm{out}}\in\mathbb{R}^{M\times N}\) is the trainable readout weight matrix,  
and \(\mathbf{c}\in\mathbb{R}^{M}\) is an optional bias term. Unlike traditional Recurrent Neural Networks (RNNs) that often struggle with vanishing or exploding gradients when learning long-term dependencies, RC models sidestep this issue by training only a linear readout layer. This simplifies the training process, making RC more stable and efficient in learning the repeating structures characteristic of periodic data.

In our study, we focus on the Echo State Network (ESN), a widely used framework of reservoir computing. Here, an important index that is closely related to the properties of ESN is the (short-term) \textit{memory capacity} (MC).
The memory capacity quantifies the network's ability to reconstruct past information from its reservoir at the network's output by computing correlations \cite{jaeger2001short,farkavs2016computational,baranvcok2014memory}. It is defined by the following equation:
\begin{equation}\label{eq:mc}
 \mathrm{MC}=\sum_{k=1}^{k_{\max }} \mathrm{MC}_{k}=\sum_{k=1}^{k_{\max }} \frac{\operatorname{cov}^{2}\left(\mathbf{u}(t-k), \mathbf{y}_{k}(t)\right)}{\operatorname{var}(\mathbf{u}(t)) \cdot \operatorname{var}\left(\mathbf{y}_{k}(t)\right)}
\end{equation}
Here, $\operatorname{cov}$ represents the covariance between the two time series, and $\operatorname{var}$ denotes the variance. $\mathbf{u}(t-k)$ refers to the input presented $k$ steps before the current input, while $\mathbf{y}_k(t)=\mathbf{\tilde{u}}(t-k)$ is its reconstructed output from the network using a linear readout. Thus, memory in ESNs is conceptualized as the network's capacity to retrieve historical information (for various $k$ values) from the reservoir through linear combinations of internal unit activations. The study \cite{baranvcok2014memory} shows that the memory capacity of ESNs depends on the properties of the structure and parameters of reservoirs, and may also be influenced by the data properties.

A key finding of our study is that introducing additional rings into the reservoir digraph enhances the orthogonality of its adjacency matrix, a property empirically demonstrated to correlate with improved reservoir performance. The orthogonality of reservoir adjacent matrices is closely related to the memory capability of reservoirs, which influences the prediction of reservoirs \cite{strauss2012design}.
Mathematically, orthogonality is a special case of linear independence, which ensures that the matrix columns are not only linearly independent but also mutually orthogonal. Improving the orthogonality of a reservoir matrix reduces overlap and interference between dataset signals stored at different time steps, enabling the network to better distinguish and retrieve distinct inputs. Moreover, orthogonal matrix columns or rows increase the probability of the matrix being full-rank, which is crucial for maximizing memory capacity. Additionally, enhanced orthogonality improves numerical stability, ensuring that computational errors do not accumulate and allowing the network to store and recall information with greater accuracy.
In summary, enhancing the orthogonality of reservoir matrices maximizes short-term memory capacity and improves the network's ability to distinguish and retrieve different input signals \cite{strauss2012design}. This provides a theoretical base for our optimization method, which focuses on improving the orthogonality of reservoir matrices.

Hence, we aim to increase the orthogonality of the reservoir adjacent matrices to improve their capabilities. 
We use the following measurement to measure the orthogonality of a matrix:
\begin{definition}
    Suppose $A$ is a matrix. Consider all its column vectors $\{\textbf{a}_1,\textbf{a}_2,\cdots,\textbf{a}_n\}$, here we define the \textbf{orthogonality measurement} of $A$ is
$$OM(A)= \frac{1}{C_n^2}\sum_{i,j=1;i\neq j}^n |m(\textbf{a}_i,\textbf{a}_j)|,$$
where $C_n^2=\frac{n(n-1)}{2}$ is the combinatorial number, $m(\textbf{a}_i,\textbf{a}_j)=1$ if $\textbf{a}_i$ or $\textbf{a}_j$ is zero vector, and $m(\textbf{a}_i,\textbf{a}_j)=\cos{<\frac{\textbf{a}_i}{\|\textbf{a}_i\|},\frac{\textbf{a}_j}{\|\textbf{a}_j\|}>}$ ($\|\cdot\| $ denotes the 2-norm) if both of them are nonzero vectors.
\end{definition}

It is easy to see that $OM(A)$ ranges from $[0,1]$, and the closer to zero, the more vectors in $\{\textbf{a}_1,\textbf{a}_2,\cdots,\textbf{a}_n\}$ are orthogonal to each other, which implies that the orthogonality of this matrix is greater. We clarify that this metric is a normalized adaptation of Mutual Coherence \cite{donoho2001uncertainty} and is closely related to the Frame Total Potential \cite{benedetto2003finite}. In high-dimensional dynamical systems like reservoirs, the degree of linear independence between state dimensions is critical for information processing. Our OM metric quantifies the average ``aliasing'' between basis vectors. By defining OM, we provide a quantitative tool to track how our GLMY theory-based topological optimization systematically reduces redundancy and approaches the isometry property required for optimal fading memory.

\begin{figure*}[!t]
    \centering
    \includegraphics[width=1\linewidth]{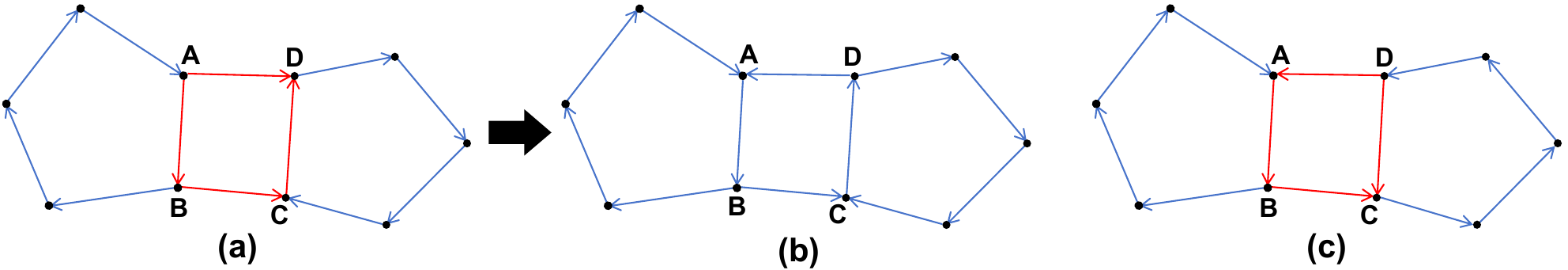}
    \caption{Illustration of compatibly modifying cycles into rings. (a) A digraph has two rings on the left and right. The middle cycle (red) is not a ring. (b) By changing the direction of $AD$, the middle cycle becomes a ring $ABCD$ compatible with the other rings. (c) The case where the middle cycle cannot be changed into a ring without destroying the other two rings, since either $AB$ or $DC$ would need to have its direction changed.}
    \label{fig:rings}
\end{figure*}

Since a reservoir can be represented by a digraph, the problem of improving the orthogonality of its adjacency matrix can also be regarded from the perspective of the reservoir digraph corresponding to the adjacency matrix. To explain this, we first introduce \textit{ring}, which is a special kind of digraph and is useful for analyzing the properties of reservoirs.
A digraph $G = (V,E)$ is a \textit{ring} if it is connected (as an undigraph), and the in-degree and out-degree of each vertex are 1. We say a digraph has rings if some of its subgraphs are rings.
According to the matrix representation of a ring, modifying a reservoir to increase the number of rings in its digraph is equivalent to making its adjacency matrix possess more fundamental cyclic submatrices, such as:
\begin{equation}\label{eq:m}
\begin{pmatrix}
 & 1 & & & \\
 & & 1 & & \\
 & & & \ddots & \\
 & & & & 1\\
 1& & & &
\end{pmatrix}
\end{equation}
or its transpose. For example, in Fig. \ref{fig:rings}, when changing from (a) to (b), the shape $\begin{tikzcd}
A \arrow[d] \arrow[r] & D           \\
B \arrow[r]           & C \arrow[u]
\end{tikzcd}$
becomes 
$\begin{tikzcd}
A \arrow[d] & D \arrow[l] \\
B \arrow[r] & C \arrow[u]
\end{tikzcd}$,
and the corresponding matrix representation changes from 
$\begin{pmatrix}
 & 1 & &1\\
 & & 1 & \\
 & & & 1\\
 & & &
\end{pmatrix}$
to
$\begin{pmatrix}
 & 1 & &\\
 & & 1 & \\
 & & & 1\\
 1& & & 
\end{pmatrix}$. Notice that the orthogonality of the adjacency matrix increases from 3 to 4. In fact, for the more general case (Equation \ref{eq:m}), since each row and column of this matrix is orthogonal to each other, the more submatrices of this type a matrix has, the greater its orthogonality will be. Combining the previous discussion of the relationship between the orthogonality of the reservoir matrix and the memory capacity of the reservoir, when we make the reservoir matrix (particularly sparse one) have more fundamental cyclic submatrices (which is equivalent to making the digraph of the reservoir have more rings), the greater the memory capacity of the reservoir is, and the more likely it is to exhibit greater predictive ability \cite{strauss2012design}.

Therefore, the core objective of this study is to enhance the orthogonality of the reservoir adjacency matrix. From a graph-theoretic perspective, this goal is equivalent to adding the number of rings in the reservoir digraph—an equivalence that forms the logical starting point for the subsequent methodological design.
To achieve this objective, the GLMY homology theory is introduced in this paper as the core analytical tool and operational framework. It is crucial to emphasize that, within the algebraic topological characterization of digraphs, rings are essentially a proper subset of the set of representative cycles contained in the GLMY homology groups. This implies that the representative cycles described by GLMY homology groups include not only rings but also other non-cyclic loop structures.

Based on the above theoretical understanding, the core idea of this study is clarified as follows: to maximize the proportion of rings within the representative cycles of GLMY homology through targeted topological operations. Specifically, it involves implementing structural transformations on those representative cycles that do not currently belong to the ring subset (i.e., non-cyclic loops) to make them become rings, thereby incorporating them into the ring subset. The essence of this process lies in systematically increasing the number of rings in the digraph by optimizing the composition of the representative cycles of GLMY homology, ultimately serving the fundamental goal of improving the orthogonality of the adjacency matrix. The specific implementation method of this transformation process will be introduced in the next subsection.

\subsection{Optimizing reservoir structure by persistent GLMY homology}
We first introduce persistent GLMY homology, which extends classical persistent homology theory to the context of weighted digraphs. Let $G = (V,E,w)$ be a weighted digraph with vertex set $V$, edge set $E$, and weight function $w: E \to \mathbb{R}_+$. For any threshold $\delta \in \mathbb{R}_+$, define the subgraph:
$$
G^{\delta} = (V, E^{\delta} = \{e \in E : w(e) \leq \delta\}).
$$
This construction yields a \textit{digraph filtration} $\{G^{\delta} \hookrightarrow G^{\delta'}\}_{\delta \leq \delta'}$, where the inclusion maps preserve the vertex set while gradually adding edges in order of increasing weight. The \textit{one-dimensional persistent GLMY homology} of $G$ is then defined as the persistent vector space:
$$
H_1 = \{H_1(G^{\delta}) \xrightarrow{i_{\delta,\delta'}} H_1(G^{\delta'})\}_{\delta \leq \delta'},
$$
where $i_{\delta,\delta'}$ denotes the induced homomorphism of the inclusion map. Here, each $G^{\delta}$ represents the state of the digraph at filtration parameter $\delta$.
This framework provides a systematic way to track the evolution of homology classes as the filtration parameter increases, revealing persistent topological features of the weighted digraph across multiple scales.

In order to find the generators of one-dimensional GLMY homology of a digraph, we first introduce some special structures.
A \textit{bigon} is a sequence of two distinct vertices $a, b \in V$ such that $a\to b, b\to a$.
A \textit{(boundary) triangle} is a sequence of three distinct vertices $a, b, c \in V$ such that $a\to b, b\to c, a\to c$:
$$\begin{tikzcd}
    & b \arrow[rd] &   \\
    a \arrow[ru] \arrow[rr] &              & c
\end{tikzcd}$$
A \textit{(boundary) square} is a sequence of four distinct vertices $a, b, c, d \in V$ such that $a\to b, b\to c, a\to d, d\to c$:
$$\begin{tikzcd}
    b \arrow[r]           & c           \\
    a \arrow[u] \arrow[r] & d \arrow[u]
\end{tikzcd}$$
Then, the one-dimensional GLMY homology of simple digraphs is clear because of the following Theorem \cite{dey2022efficient}. This theorem establishes the theoretical foundation for our operation to reservoirs.

\noindent
{\bf Theorem.}{\it ~Let $G=(V, E)$ be a simple digraph. Let $Z_1=\operatorname{Ker}\left(\left.\partial\right|_{\Omega_1}\right), B_1=\operatorname{Im}\left(\left.\partial\right|_{\Omega_{2}}\right) $ , and let $Q$ denote the space generated by all bigons and boundary triangles and boundary squares in $G$. Then, we have $B_1=Q$. Hence, the one-dimensional GLMY homology group satisfies that $H_1=Z_1 /Q$.}

Consequently, any 1-cycle in a digraph that does not form a bigon, boundary triangle, or boundary square serves as a generator of the one-dimensional GLMY homology. In our study, we employ an algorithm that computes the one-dimensional persistent GLMY homology \cite{dey2022efficient} to identify the minimal generators of $H_1(G)$ for a given digraph $G$. These minimal generators are called the \textit{minimal representative cycles}\cite{grigor2012homologies,dey2022efficient} of the $H_1(G)$.

Now we introduce our approach for optimizing reservoir structures using GLMY homology to enhance their predictive capabilities. As we have mentioned above, a key distinguishing feature of reservoirs compared to feedforward neural networks is the presence of rings in their architecture, which have been shown to possibly influence their time series prediction performance. 
Therefore, we aim to increase the number of rings in a given reservoir to optimize its structure. Specifically, when a reservoir is given, we first compute the persistent GLMY homology of its digraph and get the minimal representative cycles of one-dimensional GLMY homology groups. 
Since a ring is a special type of minimal representative cycle, we can modify other representative cycles into rings by changing some edges' directions in a compatible way (this will be introduced later).
In other words, these representative cycles are used to modify the structure of the reservoir to increase the number of rings in the reservoir digraph. In our experiments, the reservoir can get better prediction ability on three tested datasets after adding the rings in the reservoir digraph. The flowchart of our method and experimental procedures is shown in Fig. \ref{fig:flowchart}.

From a theoretical perspective, the 1-dimensional persistence diagram (1-PD) plays a crucial role in our methodology. Each point in the 1-PD corresponds to a minimal representative cycle in the digraph. Our first step is to check whether each minimal representative cycle forms a closed loop, where edges are connected end-to-end in a sequential and continuous manner, which we define as a ``ring''. This verification process is essential because only cycles with a proper ring structure can fully contribute to the reservoir's predictive capabilities.
When a minimal representative cycle fails to meet the ring condition, we employ an edge direction modification process. The primary objective of this process is to transform the non-ring representative cycle into a ring while ensuring that all existing rings within the digraph are preserved. For each individual representative cycle, we begin by attempting to modify its edges to achieve the desired ring structure. However, during the edge direction modification, if there are common edges among the current cycle and existing rings, changing the direction of these shared edges may inadvertently destroy these existing rings (see Fig. \ref{fig:rings} for an example). In such cases, where altering the current cycle would be incompatible with the preservation of existing rings, we adopt a conservative approach. Instead of forcing the modification and risking the disruption of the existing rings, we abandon the attempt to change the current cycle and proceed to the next representative cycle in the sequence.

To execute this optimization method comprehensively, we apply the above procedure sequentially to all representative cycles corresponding to the points in the 1-PD. Through this iterative process, we gradually and significantly increase the number of rings within the reservoir digraph. As the number of well-formed rings grows, the reservoir's ability to capture complex patterns improves, leading to enhanced predictive performance. This systematic approach not only leverages the theoretical insights from GLMY homology but also provides a practical and effective solution for reservoir optimization.

The above method can be summarized as the following Algorithm \ref{alg:reservoir-optimization}.

\begin{algorithm}[H]  
\renewcommand{\algorithmicrequire}{\textbf{Input:}}    
\renewcommand{\algorithmicensure}{\textbf{Return:}}    
\caption{Reservoir Optimization by GLMY Homology}  
\label{alg:reservoir-optimization}  
    \begin{algorithmic}[1]  
    	\REQUIRE A digraph $G = (V, E)$ with $n$ vertices and $m$ edges representing a reservoir.  
    	\STATE Computing the minimal representative cycles of the 1D GLMY homology of $G$. Denote these cycles by $C$.
    	\STATE \textbf{For} each cycle in $C$:
            \STATE\hspace{1em}\textbf{If} the cycle is already a ring:
            \STATE\hspace{2em} Turn to the next cycle.
            \STATE\hspace{1em}\textbf{Else}
            \STATE\hspace{2em}\textbf{If} the cycle is not a ring but can be modified as a ring without destroying existing rings:
            \STATE\hspace{3em} Change some edge directions to make it a ring. 
            \STATE\hspace{2em}\textbf{Else} 
            \STATE\hspace{3em}Turn to the next cycle.
            \STATE\hspace{1em}\textbf{EndIf}
    	\STATE\textbf{EndFor}
    	\ENSURE The optimized digraph $G$ of the reservoir. 
    \end{algorithmic}  
\end{algorithm} 

Now we analyze the complexity of Algorithm \ref{alg:reservoir-optimization}. According to the theoretical bounds provided by \cite{dey2022efficient}, computing 1-dimensional persistent GLMY homology and extracting the minimal representative cycles (the first step of our method) follows the complexity of $O(r m^{\omega-1} + m^\omega + n m g^{\omega-1})$, here $$r=\min \{ a(G)m, \sum_{(u,v)\in E}(d_{in}(u)+d_{out}(v)) \},$$
$a(G) = O(n)$ is the so-called arboricity of $G$, $\omega < 2.373$ is the exponent for matrix multiplication, and $g=rank(H_1)$.
Upon obtaining the set of $g$ minimal representative cycles, the algorithm performs a targeted structural refinement. This phase is characterized by its linear efficiency relative to the reservoir's sparsity: For each identified generator, the algorithm verifies its cyclic structure and performs a compatible edge-direction modification. The complexity for each generator is $O(L_i)$, where $L_i$ is the length of the cycle. Summed over all $g$ generators, the total complexity for this stage is $O(\sum_{i=1}^{g} L_i)$. In practice, often empirically near-linear in $m$, the number of edges in the sparse reservoir. Consequently, under the sparse setting considered in our experiments, the complexity of Algorithm \ref{alg:reservoir-optimization} is $O(r m^{\omega-1} + m^\omega + n m g^{\omega-1})$, concentrated in the initial topology computation.

\section{Experiments and Discussions}\label{sec:exp}
In this section, we introduce the datasets used and their topology analysis, and illustrate our experimental results. Our experiments aim to rigorously evaluate prediction performance before and after implementing the optimization process on reservoir digraphs. We will also give some discussions about the proposed method.

In the following subsections, we will use the following designs and parameter settings. In our experimental setup, we assumed that there is at most one edge between every two nodes, and all edges share identical weights for simplicity. 
The spectral radius \footnote{The spectral radius of a square matrix
\(A\) is defined as \(\rho(A)=\max_{i}|\lambda_{i}|\), where
\(\{\lambda_{i}\}\) are the eigenvalues of \(A\).} of the reservoir in our experiments is set as $\lambda_{target} = 0.8$ since empirical studies~\cite{venayagamoorthy2009effects} indicate that a spectral radius close to \(0.8\) realizes a favourable trade‑off between fading memory and nonlinear amplification, thereby yielding best performance for the reservoir. The setting of the spectral radius can be achieved by the following steps: firstly, for a randomly generated reservoir, we assign each edge an identity weight, for example, one, to get the initial reservoir digraph $W_{initial}$. Then we can compute the maximum eigenvalue $\lambda_{max}$ of the reservoir digraph, followed by adjusting the spectral radius of the reservoir to $\lambda_{target}$ using the following formula:
$$W_{r}=\lambda_{target} \cdot \frac{W_{initial}}{\lambda_{max}}.$$

\begin{figure*}[p]
    \centering
    \includegraphics[width=1.0\linewidth]{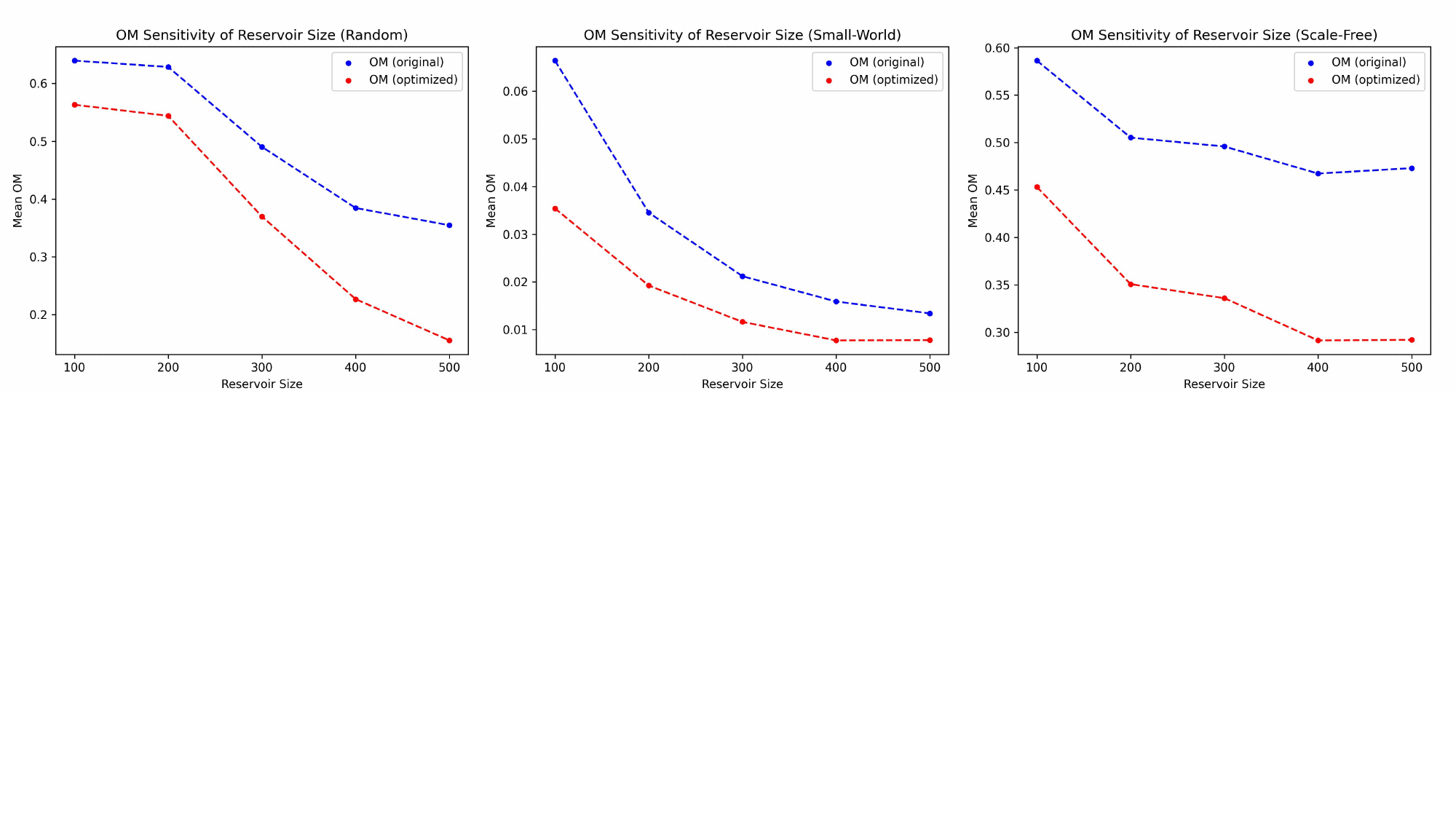}
    \caption{Reservoir size sensitivity analysis. Here we analyze reservoir sizes ranging from 100 to 500 and calculate the orthogonality measurement (OM) before and after optimization. The results presented from left to right correspond to Random, Small-world, and Scale-free initialization, respectively.}
    \label{fig:size}
\end{figure*}

\begin{figure*}[p]
    \centering
    \includegraphics[width=1.0\linewidth]{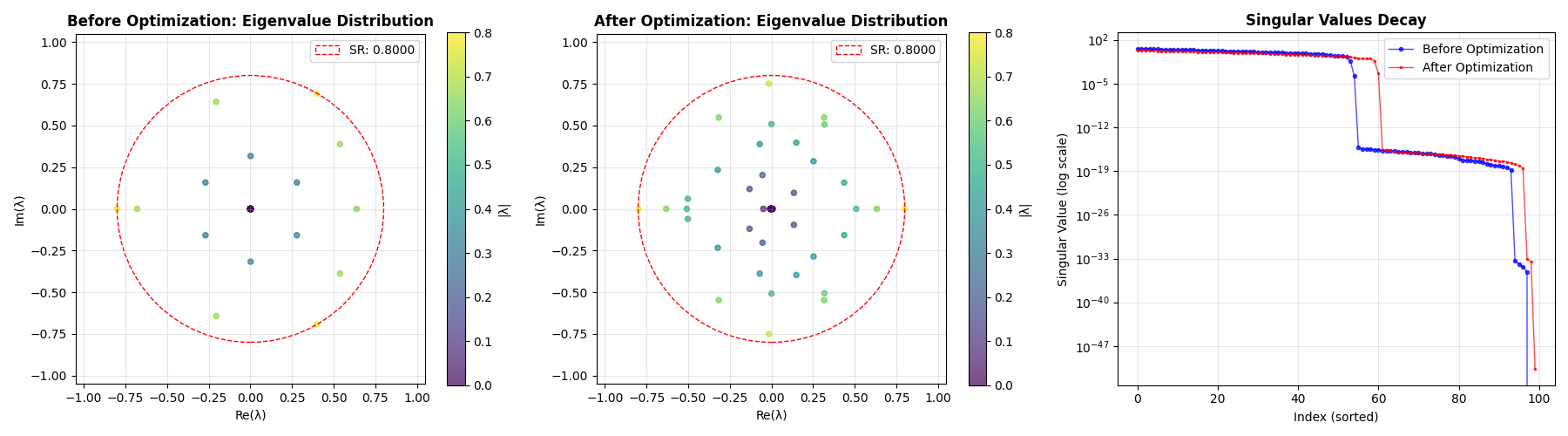}
    \caption{Spectral analysis of a reservoir with 100 nodes, 0.99 sparsity, and a fixed spectral radius of 0.8 before and after optimization. Left: eigenvalue distribution before optimization. Middle: eigenvalue distribution after optimization. Right: singular values before (blue) and after (red) optimization.}
    \label{fig:spec}
\end{figure*}

\begin{figure*}[p]
    \centering
    \includegraphics[width=1.0\linewidth]{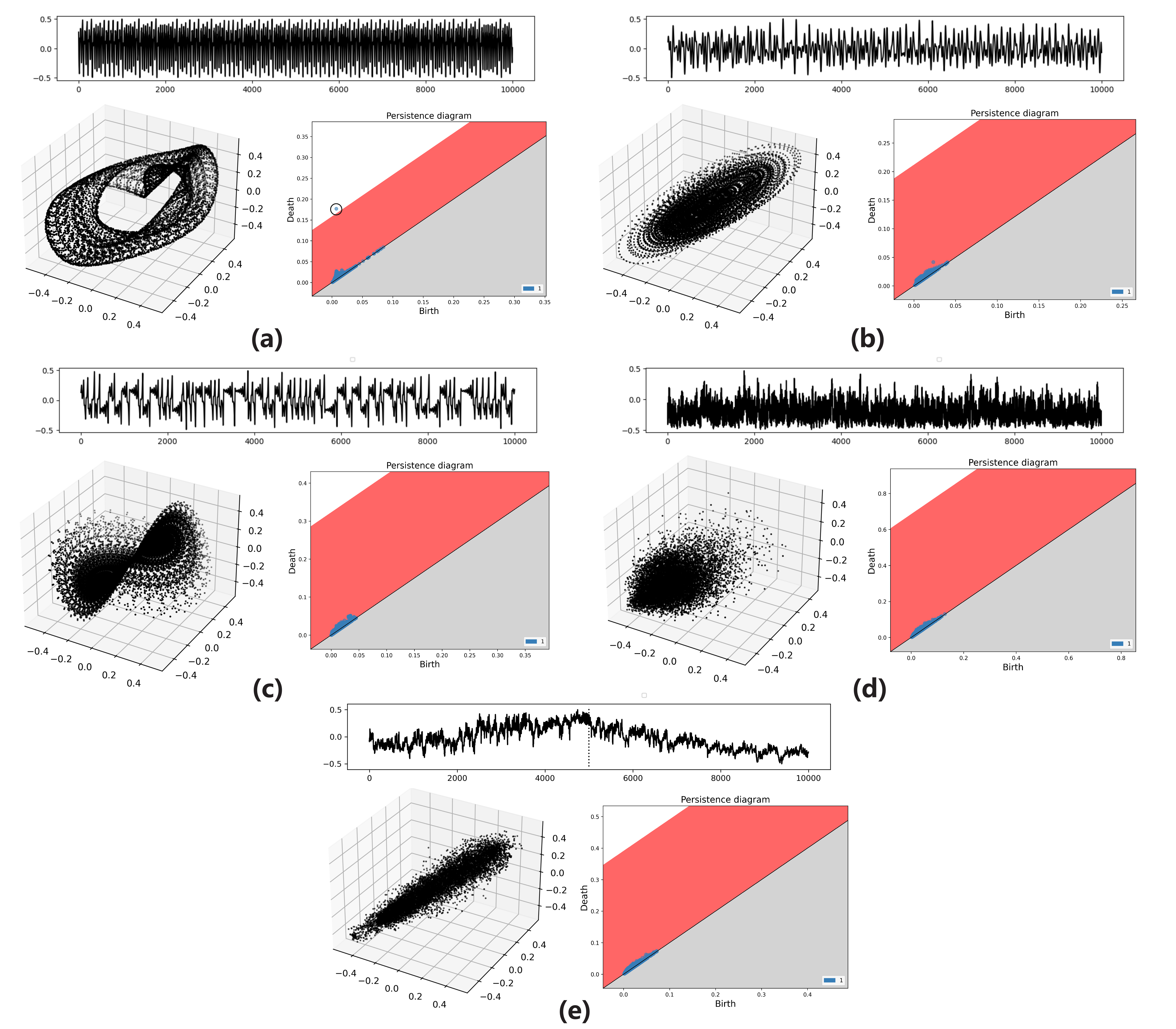}
    \caption{Illustration of five datasets, and their Time-delay embedding results in $\mathbb{R}^3$ and the one-dimensional persistence diagrams. (a) Result of Mackey-Glass Data. Periodicity Index is 0.1696. (b) Result of MSO Data. Periodicity Index is 0.0183. (c) Result of Lorenz System Data. Periodicity Index is 0.0183. (d) Result of NARMA Data. Periodicity Index is 0.0269. (e) Result of ETT Data. Periodicity Index is 0.0135. }
    \label{fig:PDs}
\end{figure*}

\subsection{Reservoir Size Sensitivity and Reservoir Dynamics Analysis}
First, we examine the sensitivity of our method with respect to reservoir size. In particular, we focus on optimizing the orthogonality measurement (OM) for varying reservoir sizes (ranging from 100 nodes to 500 nodes) and initialization strategies (Random, Small-world, and Scale-free). Fig. \ref{fig:size} demonstrates that the proposed optimization consistently reduces OM across reservoir sizes and initialization types, with the largest absolute improvements observed for initially less-orthogonal (random and scale-free) reservoirs, while preserving the spectral-radius constraint. It is noteworthy that these results are obtained while maintaining the same spectral radius constraint ($\lambda=0.8$), thus supporting our assertion that orthogonality can be improved without sacrificing stability by adjusting the structure of rings.

Then, we illustrate how our optimization method affects the reservoir dynamics. Thanks to the size sensitivity analysis, it suffices to analyze a specific reservoir size. Here we conducted a spectral analysis on a sparse reservoir with 100 nodes, 0.99 sparsity, and a fixed spectral radius of 0.8, to isolate the structural effects. Fig. \ref{fig:spec} shows the eigenvalue distributions and singular value decay before and after optimization while keeping the spectral radius fixed at 0.8. After optimization, the eigenvalues did not exhibit the situation where a few extremely slow modes were pushed towards the spectral radius boundary; instead, the modulus distribution became more uniform. This indicates that the optimization does not rely solely on extremely slow modes but instead enhances the overall contribution of multiple moderately decaying modes, thereby giving the state space a higher effective representational capability in multiple directions and improving memory capacity. Moreover, the post-optimization singular value spectrum is noticeably flatter in its mid-range and exhibits fewer near-zero singular values, indicating a more uniform energy distribution across state directions. Dynamically, this corresponds to a redistribution of modal contributions: more moderate-decay modes rather than a few dominant very-slow modes, which yields a richer set of approximately orthogonal state components for the linear readout to exploit.

Additionally, in our spectral analysis, the eigenvalues of the reservoir weight matrix remain strictly within the target spectral radius both before and after optimization, which indicates that the optimization does not compromise this fundamental stability and hence preserves the fading memory property. Moreover, the post-optimization eigenvalue and singular value distributions suggest a redistribution of modal contributions without inducing a few excessively slow modes. Instead, a broader set of moderate-decay modes emerges, enriching the reservoir's state representation under stable dynamics. Such a shift enhances the reservoir's effective memory capacity while maintaining controlled fading of older inputs, consistent with the theoretical notion of fading memory.

\subsection{Datasets and their topology}
To validate our approach and compare results from different datasets on the same scale and for the convenience of comparing topological difference, we employ the following datasets with value normalized into the range $[-0.5,0.5]$:
\begin{enumerate}
    \item \textbf{Mackey-Glass System}: This classical chaotic system evaluates time series prediction models through the discrete approximation:
    $$
        y(t+1) = y(t) + \delta\left(a\frac{y(t-\tau/\delta)}{1+y(t-\tau/\delta)^n} - by(t)\right),
    $$
    with standard parameters $\delta=0.1$, $a=0.2$, $b=-0.1$, $n=10$, and chaotic regime $\tau=17$. The task involves $k$-step prediction using historical data $u(1,\dots,t)$ to forecast $u(t+k)$. In our experiments, we use the first 5000 steps for training and the following 5000 steps for prediction.

    \item \textbf{Multiple Superimposed Oscillator series}: The Multiple Superimposed Oscillator (MSO) series prediction is also an important problem. The core concept behind the MSO is to combine multiple oscillator signals into a single composite indicator by superimposing and normalizing them. It can be described by the following equation:
    $$y(t)=\sum_{i=1}^m\operatorname{sin}(\alpha_i t),$$
    where $m$ is the number of summed sine waves. To test ESN, a common way is to set $m=5$ and $\alpha_1=0.2, \alpha_2=0.311, \alpha_3=0.42, \alpha_4=0.51, \alpha_5=0.63.$

    \item \textbf{Lorenz System}: A typical benchmark for chaotic systems:
    $$
        \begin{cases}
        \dot{x} = -a x(t) + a y(t) \\
        \dot{y} = b x(t) - y(t) - x(t) z(t) \\
        \dot{z} = x(t) y(t) - c z(t)
        \end{cases}
    $$
    with $a=10/4$, $b=28$, $c=8/3$. The Runge-Kutta method can generate a sample of length $L$ from the initial conditions $(x_0,y_0,t_0)$ with a step size of 0.01. In our experiments, the $y$-axis is chosen for our prediction task, and we also use the first 5000 steps for training and the following 5000 steps for prediction. The first 1000 input samples were discarded after being fed into the network.

    \item \textbf{NARMA System}: This nonlinear auto-regressive moving average system presents a challenging identification problem:
    \begin{align*}   
        y(t+1) = 0.3y(t) +& 0.05y(t)\sum_{i=0}^9 y(t-i) + \\&1.5u(t-9)u(t) + 0.1,
    \end{align*}
    where $u(t)\sim\mathcal{U}[0,0.5]$ (uniform distribution). Initialized with $y(t)=0$ for $t=1,\dots,10$, it tests echo state networks through one-step prediction. In our experiments, we use the first 5000 steps for training and the following 5000 steps for prediction. The first 500 input samples were discarded after being fed into the network.

    \item \textbf{Electricity Transformer Dataset}:
    The Electricity Transformer Dataset (ETT) dataset is one of the widely used standard public datasets in the field of long-series time series forecasting in recent years, published and systematically organized by Zhou et al. \cite{haoyietal-informer-2021}. This dataset records various monitoring variables such as load power and oil temperature of transformers in power systems during long-term operation. It exhibits significant seasonal and periodic components, while also containing random fluctuations, making it suitable as a forecasting benchmark for real-world industrial scenarios. We use an hourly subset of the ETT dataset, selecting 10,000 hours of monitoring data beginning on 2017/1/1. In the experiments, transformer oil temperature is used as the predicting target. We use the first 5000 steps for training and the following 5000 steps for prediction.

\end{enumerate}

To analyze the periodic differences between datasets, we employ \textit{Time-Delay Embedding} (TDE) \cite{takens2006detecting} to embed the time series into higher-dimensional space, followed by computing their one-dimensional persistence diagrams (1-PDs) from constructing the alpha filtration \cite{edelsbrunner2010computational} of the embedded point clouds.
Time-Delay Embedding is a powerful technique used in the analysis of time series data, which transforms a single observed time series into a point cloud in high-dimensional space. This method leverages the concept of phase space reconstruction, allowing for the visualization and analysis of complex time series from limited measurements \cite{takens2006detecting}.

Given a scalar time series $\{x_t\}_{t=1}^{N}$, where $N$ is the number of observations, the TDE process constructs a set of points in a higher-dimensional space (called the embedding space) using a specified embedding dimension $d$ and a time delay parameter $\tau$. Each point in this new representation is formed as follows:
$$
\mathbf{X}_i = [x_i, x_{i+\tau}, x_{i+2\tau}, \ldots, x_{i+(d-1)\tau}]^T,
$$
where $i = 1, 2, \ldots, N-(d-1)\tau$, ensuring that all components of each $\mathbf{X}_i$ are within the bounds of the original time series. 

The choice of the embedding dimension $d$ and the time delay $\tau$ plays a crucial role in the effectiveness of TDE. The embedding dimension determines the number of past values included in each vector, capturing the dynamics of the underlying system. Meanwhile, the time delay specifies how far apart these values are spaced in time, influencing the independence between the components of the embedded vectors. In our study, by testing a variety of parameter selections, we embed the aforementioned time series into $\mathbb{R}^3$ with a delay parameter of $\tau=5$, which gives a clear and intuitive illustration of the topological characteristics of the embedded data.
This transformation not only enhances the estimation precision of models but also reveals hidden structures and patterns in the data that are not apparent in the raw time series form.
The one-dimensional persistence diagram captures the topological features of loops present in the point cloud. When a point in the 1-PD exhibits significantly greater persistence than others, this indicates both the existence of a prominent loop structure in the embedded point cloud and a strong periodic mode in the original time series \cite{perea2015sliding}. This correspondence between persistent homology features and time series periodicity provides valuable insight into the underlying dynamics of the system. Therefore, we can use the Max Persistence in a 1-PD as a Periodicity Index (PI) to detect the periodicity of a time series:
$$PI = \max{ \{d_i - b_i \:|\: (b_i,d_i)\in PD_1 \}}.$$
In this representation, the periodic or quasi-periodic dynamics of the signal manifest as cyclic attractors \cite{takens2006detecting,perea2015sliding}.
A higher $PI$ indicates a more robust cyclic attractor in the data's underlying dynamics, representing stronger periodicity.
The results are illustrated in Fig. \ref{fig:PDs}, where we illustrate the original time series, the embedded point clouds in $\mathbb{R}^3$, and the 1D persistence diagram with confidence sets. The title of Fig. \ref{fig:PDs} records the $PI$ of these datasets. These results reveal distinct topological characteristics: the Mackey-Glass data exhibits a significant 1-dimensional topological feature as its 1-PD has one point located above the red area, while others show no significant 1-dimensional topological structures. Hence, these five datasets differ in their prominence of periodicity.

It is widely observed that Reservoir Computing models may demonstrate superior performance when dealing with datasets exhibiting pronounced periodicity \cite{jaeger2001echo}. This enhanced capability stems from RC's unique architectural features, which are well-suited to capture and exploit the complex temporal dependencies inherent in periodic signals. The fixed, randomly connected ``reservoir'' acts as a dynamic, high-dimensional feature extractor, mapping input sequences into a rich, non-linear state space. This intrinsic non-linearity and recurrent connectivity enable the reservoir to effectively represent intricate periodic patterns and their underlying dynamics, even when such patterns are embedded within noise or non-stationary components. Furthermore, the inherent memory capacity of the reservoir allows it to retain information about past inputs over extended periods, which is crucial for learning and predicting long-term periodic cycles. According to the operation of TDE, the topology of the embedded data relates to its periodicity, hence influencing the prediction results of the reservoir. This will be shown later in the experimental results.

\subsection{Initial reservoir generation and its optimization}
In our experiments, we tested three types of initial construction strategies for the reservoir. The first one is constructing the reservoir randomly, which typically results in a sparse reservoir digraph matrix. Additionally, we also test the small-world reservoir constructed with the Watts–Strogatz model \cite{watts1998collective} and the scale-free reservoir produced by the Barabási–Albert (BA) model \cite{barabasi1999emergence}. These manually designed network structures have been shown in related studies to provide superior prediction accuracy and a wider spectral radius, which provide a diverse set of benchmarks for our optimization method to validate its effectiveness.

Here, we compute the orthogonality measurement (OM) defined in section \ref{sec:4.1} of each initialization method, comparing their digraph matrices' OM before and after optimization. Additionally, the number of rings that are added to the digraph is also counted. This evaluation is crucial because research suggests that an increase in the orthogonality of the reservoir matrix can significantly enhance its short-term memory capacity and improve the network's ability to distinguish and retrieve different input signals. From a topological perspective, increasing the number of rings in the reservoir digraph is equivalent to its adjacency matrix having more fundamental cycle submatrices, which is the key to improving orthogonality.

In this computation, the node number of the reservoir digraph is 500, which means the digraph matrix has the size $500\times 500$. Hence, the orthogonality of each digraph matrix will not be greater than 500. In the Random configuration, we set the sparsity of the reservoir to 0.99, which means that only $1\%$ of the possible edges in the fully connected graph of all nodes, which are randomly selected, actually exist. Due to the presence of a large number of zero elements, the initial orthogonality of the matrix will not be too large. In the small-world configuration, each node was initially connected to its $k=10$ nearest neighbors on a ring lattice, after which every edge was rewired with probability $p=0.3$. In the scale-free configuration, each newly added node established $m=8$ directed links according to preferential attachment. In the following discussion, we perform 10 tests for each of the initialization methods and datasets and take the overall average of each test index, which illustrates the general applicability of our method. We compute the persistent GLMY homology and minimal representative cycles using the code provided by \cite{dey2022efficient}. Again, we note that there is at most one edge between any two nodes, and all edges have identical weights for simplification.

Table \ref{tab:comparison} illustrates the comparison of the average orthogonality measurement (OM) and ring addition number of the three different initialization methods, as well as the average optimization running time. 
For the Random initialization approach, the reservoir's initial sparsity results in a significant number of zero columns. Our optimization method effectively addresses this by substantially improving the average level of orthogonality. Additionally, our approach introduces numerous cycles into the reservoir digraph, strongly demonstrating that the method, based on GLMY homology theory, can indeed optimize the reservoir's structure. Similar results were observed in the Small-world initialization cases; although the increase in cycles was less than in the Random cases, the average OM was still reduced by approximately half. In the Scale-free approach, our optimization also reduced its OM significantly.
These results, while showing variations in the original orthogonality and the number of cycles that can be added across different initialization methods, collectively validate our approach's effectiveness in increasing the orthogonality of the matrices. We can also infer that the orthogonality enhancement will be more significant with larger reservoir matrices.

In the following subsections, we will compute the memory capacity to further validate this observation and provide a more comprehensive performance evaluation. Moreover, we will also provide some related discussions later.

\begin{table}[]
    \centering
    \caption{Comparison of average orthogonality measurement (OM), ring addition, and optimization running time across different initialization methods for $500\times 500$ reservoir digraph matrices.}
    \begin{tabular}{cccc}
        \hline
        & Random & Small-world & Scale-free \\
        \hline
        Original OM  & 0.3227 & 0.0134 & 0.4795 \\
        \hline
        Optimized OM & 0.1397 & 0.0074 & 0.2909 \\
        \hline
        Added rings  & 112 & 97 & 88 \\
        \hline
        Running time  & 762.5s & 540.5s & 347.5s \\
        \hline
    \end{tabular}
    \label{tab:comparison}
\end{table}

\begin{figure*}[p]
    \centering
    \includegraphics[width=0.95\linewidth]{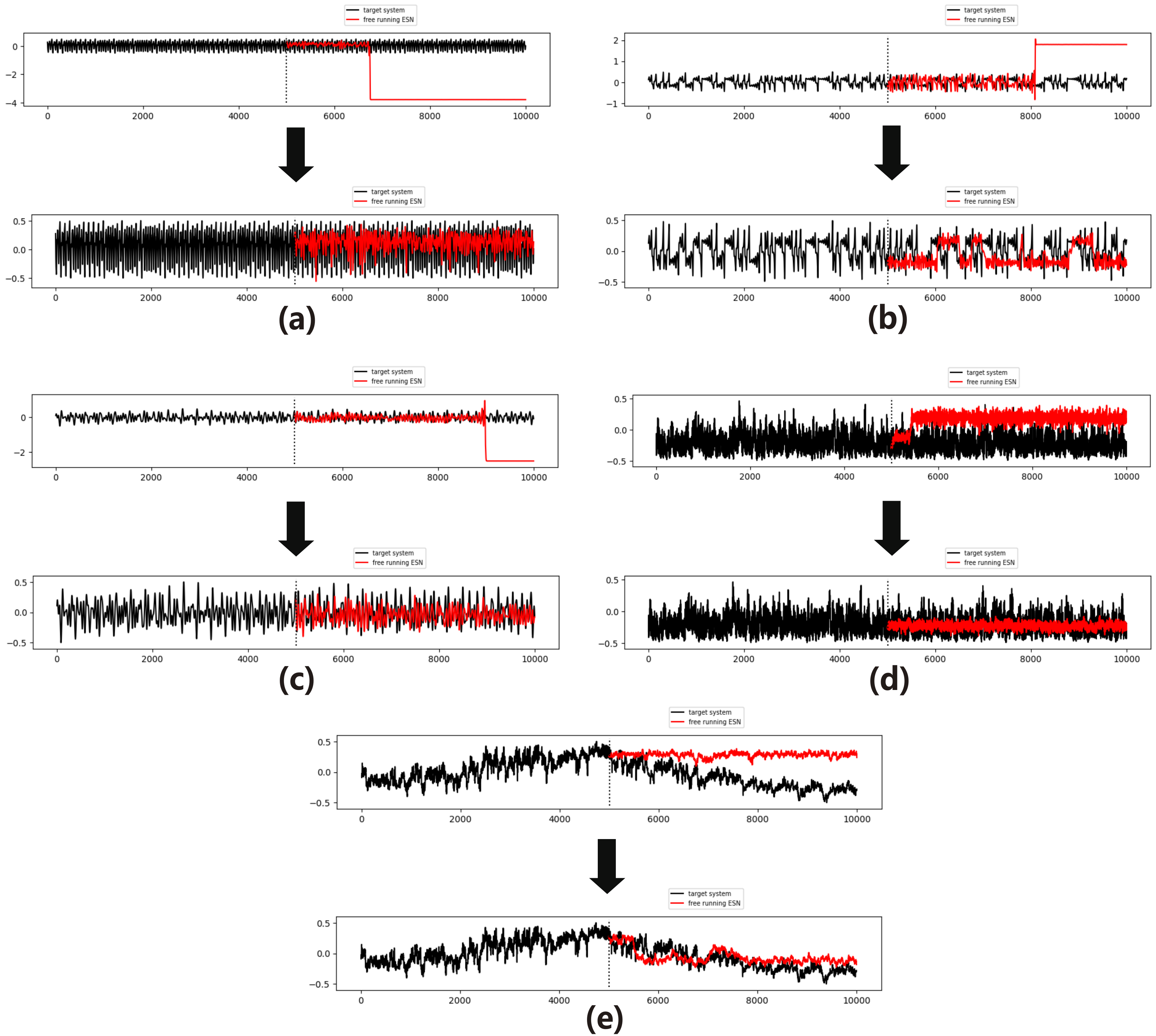}
    \caption{Illustration of the prediction results on five datasets. (a)--(e) are results of the Mackey-Glass, MSO, Lorenz system, NARMA, and ETT datasets, respectively. Figures on the top are prediction results of the initial randomly generated reservoir, with black being the real data and red being the predicted values. It can be seen that after a period of time, the reservoir loses its predictive ability; Figures on the bottom are the prediction results of the optimized reservoir, whose predictive ability has been significantly enhanced. }
    \label{fig:datavisual}
\end{figure*}

\begin{table*}[]
    \centering
    \caption{Memory Capacity (MC) (Mean$\pm$Std) tested on five datasets before and after GLMY theory-based optimization. Here we compared three initialization approaches (Random, Small-world, Scale-free).}
    \resizebox{0.55\linewidth}{!}{
    \begin{tabular}{cccc}
        \hline
        \hline
        \multicolumn{4}{c}{\textbf{Mackey-Glass dataset}}\\
        \hline
        & Random & Small-world & Scale-free \\
        \hline
        Origin & $1.6497_{3.3122}$ & $4.6585_{6.4405}$ & $4.6907_{10.8505}$\\
        \hline
        \textbf{GLMY Optimized} & $13.0021_{13.9663}$ & $14.6279_{8.1599}$ & $14.5722_{11.8339}$\\
        \hline
        \hline
        \multicolumn{4}{c}{\textbf{MSO dataset}}\\
        \hline
        & Random & Small-world & Scale-free \\
        \hline
        Origin & $3.0975_{2.9030}$ & $2.4869_{1.0932}$ & $1.8996_{1.8761}$\\
        \hline
        \textbf{GLMY Optimized} & $3.4561_{2.2308}$ & $2.7297_{1.4397}$ & $2.7929_{1.2467}$\\
        \hline
        \hline
        \multicolumn{4}{c}{\textbf{Lorenz system dataset}}\\
        \hline
        & Random & Small-world & Scale-free \\
        \hline
        Origin & $2.7488_{2.2589}$ & $2.8108_{2.3106}$ & $2.0317_{0.8796}$ \\
        \hline
        \textbf{GLMY Optimized} & $5.2335_{3.8498}$ & $3.7102_{2.0723}$ & $3.9387_{2.5507}$\\
        \hline
        \hline
        \multicolumn{4}{c}{\textbf{NARMA dataset}}\\
        \hline
        & Random & Small-world & Scale-free \\
        \hline
        Origin & $0.6518_{0.1461}$ & $0.5548_{0.2050}$ & $0.6869_{0.3268}$\\
        \hline
        \textbf{GLMY Optimized} & $0.6923_{0.2479}$ & $0.7996_{0.2332}$ & $0.7227_{0.2506}$\\
        \hline
        \hline
        \multicolumn{4}{c}{\textbf{ETT dataset}}\\
        \hline
        & Random & Small-world & Scale-free \\
        \hline
        Origin & $18.0664_{21.5056}$ & $7.8563_{9.2927}$ & $5.3869_{62.1683}$\\
        \hline
        \textbf{GLMY Optimized} & $51.0430_{79.0727}$ & $31.4610_{4.9896}$ & $21.7803_{18.5071}$\\
        \hline
        \hline
    \end{tabular}
    }
    \label{tab:mc}
\end{table*}

\subsection{Memory capacity analysis}
In the subsequent experiments, the reservoir digraph's node number is still set as 500. For the Random initial method, the reservoir sparsity is set to 0.99 (i.e., $1\%$ connectivity), which is a typical setting in practice \cite{jaeger2004harnessing}. This configuration results in a large and sparse reservoir, which effectively leverages the greater predictive power of reservoir computing.

Now we begin introducing the results of the average memory capacity (MC) of reservoirs initialized by the three aforementioned methods, both before and after optimization, which demonstrates the improvement achieved by our approach. Here, the $k_{max}$ in Equation \ref{eq:mc} was set to $1.4n$ (where $n$ denotes the reservoir size), following the experimental setup described in \cite{farkavs2016computational}. For illustrative purposes, here we test the three initialization methods on the five datasets mentioned before, and compute the average MC and standard deviation of each tested dataset. The results presented in Table \ref{tab:mc} clearly indicate that our method enhances the MC of the reservoirs. Although the initial MC values vary across different initialization methods due to their distinct reservoir architectures, our optimization approach exhibits a consistent effect on each method, leading to improvements in MC across the board. Moreover, it can be concluded that the improvement of the MC is related to the type of dataset and the initialization approach. 

The optimization method consistently and effectively enhances the MC across all tested datasets and initialization schemes. However, the magnitude of this improvement varies significantly. For datasets like Mackey-Glass and ETT, the optimized reservoir shows a dramatic increase in MC, suggesting that the optimization is highly effective for these specific types of time series data. In contrast, for the other datasets, the optimization provides a more modest boost to MC. Furthermore, the choice of initialization method plays a crucial role. These results are consistent with our earlier introduction in section \ref{sec:4.1} that MC depends on the properties of the structure and parameters of reservoirs, and may also be influenced by the data properties.

To further visualize these improvements, we showcase some typical experimental results from our study, as depicted in Fig. \ref{fig:datavisual}. After optimization, the duration for which the prediction results remain close to the real dataset is significantly extended compared to the pre-optimization state. This extension not only confirms the observable improvement in prediction length but also quantifies its substantial nature, which can be directly attributed to the enhanced memory capacity of the optimized reservoir. Importantly, this phenomenon was consistently observed across lots of the experimental cases, validating its generality.

Therefore, these results demonstrate that after optimization, the reservoirs exhibit enhanced MC, enabling them to better capture the intrinsic features of the datasets. This improvement in feature capture capability directly translates to a significant enhancement in prediction performance. These results illustrate the effectiveness of our optimization method in boosting reservoir computing performance through memory capacity enhancement.

\begin{table*}[]
    \centering
    \caption{RMSE (Mean$\pm$Std) tested on five datasets before and after GLMY theory-based optimization, including a Random Rewiring baseline. Here we compared three initialization approaches (Random, Small-world, Scale-free).}
    \resizebox{0.55\linewidth}{!}{
    \begin{tabular}{cccc}
        \hline
        \hline
        \multicolumn{4}{c}{\textbf{Mackey-Glass dataset}}\\
        \hline
        & Random & Small-world & Scale-free \\
        \hline
        Origin & $2.9505_{1.8449}$ & $1.4882_{1.3749}$ & $1.7773_{0.9896}$\\
        \hline
        Random Rewiring &$3.6186_{1.3559}$ & $2.3224_{1.8731}$ & $3.4308_{1.4233}$ \\
        \hline
        \textbf{GLMY Optimized} & $0.3039_{0.0228}$ & $0.3216_{0.0292}$ & $0.3288_{0.0298}$\\
        \hline
        \hline
        \multicolumn{4}{c}{\textbf{MSO dataset}}\\
        \hline
        & Random & Small-world & Scale-free \\
        \hline
        Origin & $0.6433_{0.8226}$ & $0.7405_{1.0915}$ & $0.7579_{0.8848}$\\
        \hline
        Random Rewiring &$0.2611_{0.0855}$ & $0.3496_{0.3952}$ & $0.2323_{0.0119}$ \\
        \hline
        \textbf{GLMY Optimized} & $0.2248_{0.0102}$ & $0.2328_{0.0164}$ & $0.2216_{0.0081}$\\
        \hline
        \hline
        \multicolumn{4}{c}{\textbf{Lorenz system dataset}}\\
        \hline
        & Random & Small-world & Scale-free \\
        \hline
        Origin & $1.0598_{1.2836}$ & $0.9786_{1.0647}$ & $0.8555_{1.0417}$\\
        \hline
        Random Rewiring &$1.0706_{1.3901}$ & $0.4393_{0.6146}$ & $0.3919_{0.4582}$ \\
        \hline
        \textbf{GLMY Optimized} & $0.2474_{0.0075}$ & $0.2454_{0.0088}$ & $0.2489_{0.0088}$\\
        \hline
        \hline
        \multicolumn{4}{c}{\textbf{NARMA dataset}}\\
        \hline
        & Random & Small-world & Scale-free \\
        \hline
        Origin & $0.1515_{0.0082}$ & $0.1580_{0.0141}$ & $0.1560_{0.0178}$\\
        \hline
        Random Rewiring &$0.1539_{0.0056}$ & $0.1489_{0.0028}$ & $0.1530_{0.0060}$ \\
        \hline
        \textbf{GLMY Optimized} & $0.1487_{0.0028}$ & $0.1475_{0.0024}$ & $0.1487_{0.0035}$\\
        \hline
        \hline
        \multicolumn{4}{c}{\textbf{ETT dataset}}\\
        \hline
        & Random & Small-world & Scale-free \\
        \hline
        Origin & $0.4519_{0.0665}$ & $0.4438_{0.0358}$ & $0.4615_{0.0362}$\\
        \hline
        Random Rewiring &$0.3071_{0.1019}$ & $0.1973_{0.0173}$ & $0.3465_{0.0849}$ \\
        \hline
        \textbf{GLMY Optimized} & $0.1961_{0.0213}$ & $0.1927_{0.0165}$ & $0.2017_{0.0298}$\\
        \hline
        \hline
    \end{tabular}
    }
    \label{tab:results}
\end{table*}

\subsection{Prediction performance analysis}
Then we compute the average RMSE and standard deviation tested on five datasets before and after optimization using our approach, illustrating the improvement in the performance of reservoirs. 
The RMSE is computed as 
$$
RMSE=\sqrt{\frac1n \sum_{i=1}^n (y_i-\hat{y}_i)^2},
$$
where $n$ is the time steps to predict, $y_i$ and $\hat{y}_i$ represent the predict value and real value of the $i$-step respectively.
Table \ref{tab:time} shows the train and test time on the five datasets \footnote{All the experiments were conducted on a server equipped with dual Intel(R) Xeon(R) CPU E5-2678 v3 processors (2.50GHz, 24 cores, 48 threads) and 62 GB of RAM.}, and Table \ref{tab:results} summarizes our experimental results, which shows the tested results on five datasets using three different initial construction. Here, ``origin'' represents the original reservoirs initialized by different approaches.
A degree-preserving random rewiring baseline (``Random Rewiring'' in the Table) was implemented using repeated edge swaps, ensuring identical in-degree and out-degree distributions. The swapping time is set to 0.3 times the number of edges. The spectral radius is re-normalized to 0.8 after rewiring, consistent with our optimization method. As reported in Table \ref{tab:results}, the random rewiring fails in many cases, which may make the RMSE become larger, while ours can reduce RMSE in all three initialization methods and five datasets.
As for our optimization method, for the Random initialization, the initial reservoir digraphs are sparse and hence show weak memory capacities and prediction abilities at first, even if the dataset has obvious topological features. However, our experimental results demonstrate that even highly sparse reservoirs can achieve substantially improved prediction performance through adding the number of rings within their digraph architecture. And for the small-world and scale-free methods, since in some studies they have been proven to be effective initialization ways of reservoirs \cite{deng2006complex,kawai2019small,cui2012architecture}, we find that they have already achieve relative good performance of prediction on some datasets. But for these cases, we have also observed that after adding the number of rings, if the original reservoir already possessed great prediction capability (such as the small-world and scale-free method on the NARMA datasets), the modified results can maintain such good capability, i.e., obtaining similar RMSE or lower RMSE. On the contrary, for reservoirs with initially poor prediction capabilities (such as the Mackey-Glass dataset), the optimization process proves to be even more useful, where the modified reservoirs experience a significant enhancement in their prediction power. 

\begin{table*}[]
    \caption{Comparison of training and testing time (s) across datasets and initialization types.}
    \centering
    \begin{tabular}{ccccccc}
        \hline
        \textbf{Dataset} & \multicolumn{2}{c}{\textbf{Random}} & \multicolumn{2}{c}{\textbf{Small-world}} & \multicolumn{2}{c}{\textbf{Scale-free}} \\
        & \textbf{Train} & \textbf{Test} & \textbf{Train} & \textbf{Test} & \textbf{Train} & \textbf{Test} \\
        \hline
        Mackey-Glass & 1.8 & 0.6 & 1.2 & 0.6 & 1.7 & 0.8 \\
        MSO & 1.4 & 0.5 & 1.7 & 0.5 & 1.6 & 0.6 \\
        Lorenz System & 1.4 & 0.7 & 1.9 & 0.7 & 2.0 & 0.5 \\
        NARMA & 1.4 & 0.9 & 1.8 & 0.6 & 1.9 & 0.8 \\
        ETT & 1.2 & 0.5 & 1.9 & 0.6 & 1.8 & 0.7 \\
        \hline
    \end{tabular}
    \label{tab:time}
\end{table*}

Moreover, our optimization method demonstrates varying predictive gains across the evaluated datasets, revealing a compelling Topological Resonance effect that scales with the quantitative Periodicity Index ($PI$) (Fig. \ref{fig:PDs}). The Mackey-Glass dataset, possessing the highest $PI$ (0.1696) and prominent loop features, achieves the most dramatic RMSE reduction (a $\approx 90\%$ improvement for random initialization), suggesting that the GLMY optimization effectively aligns the reservoir's engineered cycles with the data's inherent cyclic dynamics. This sensitivity to topological persistence is further corroborated by the results on MSO, Lorenz, and NARMA. These datasets exhibit significantly lower $PI$ values (ranging from 0.0183 to 0.0269) and, as expected, show more constrained performance enhancements. Notably, an interesting observation is found in the ETT dataset. Despite its low $PI$ (0.0135), it exhibits a substantial performance gain (RMSE reduced by $\approx56\%$ for random initialization). This indicates that our method enhances reservoir performance through two distinct pathways: (1) Topological Resonance: For periodic data, the engineered cycles in the reservoir ``resonate'' with the data's cyclic attractor. (2) Memory Extension: As shown in Table 2, the GLMY optimization significantly boosts the Memory Capacity (MC) (from 18.06 to 51.04). Since the ETT task involves long-term forecasting, it benefits more from the extended fading memory provided by the ring-like structures, even in the absence of strong periodicity. This dual-pathway mechanism explains the versatility of our method across diverse datasets.

\subsection{Discussion}
\subsubsection{Comparison with Traditional Topological and Spectral Approaches}
While topological data analysis has been applied to analyze the complexity and decision boundaries of neural networks \cite{rieck2019neural,naitzat2020topology}, and spectral graph theory is widely used to ensure the stability of reservoir computing (e.g., the spectral radius condition) \cite{jaeger2001echo,lukovsevivcius2012practical}, they exhibit fundamental limitations when applied to the structural optimization of Reservoir Computing (RC). First, traditional persistent homology typically treats networks as undirected graphs or metric spaces. The resulting topological invariants, such as standard Betti numbers ($\beta_1$), quantify the number of ``holes'' or cycles without regard to edge direction. In the context of RC, this is insufficient because an undirected cycle does not necessarily imply a feedback loop, which is the physical mechanism for memory retention. A ``hole'' in the undirected sense might simply be a feedforward structure, whereas only directed cycles contribute to the recurrent dynamics essential for fading memory. GLMY homology, by contrast, is intrinsically designed for digraphs. It rigorously distinguishes between non-cyclic structures and true directed cycles, ensuring that the optimized topology directly enhances the information recurrence capabilities of the reservoir. Second, spectral graph theory, which analyzes the eigenvalues (spectra) of the Laplacian or Adjacency matrices, provides valuable insights into global properties like network connectivity and stability (e.g., the spectral radius). However, spectral metrics are global statistics; they do not explicitly identify or locate specific local structures that can be surgically modified. For instance, two reservoirs with identical spectral radii can have vastly different cycle distributions and memory capacities. The approach grounded in our GLMY theory provides a perspective that is both local and structural, enabling the accurate identification and modification of particular minimal representative cycles (generators of $H_1$) to enhance orthogonality. This degree of detail is unattainable through spectral methods.

\subsubsection{Quantitative Interplay between Dataset Topology and Performance}
The predictive performance of our GLMY theory-based optimization demonstrates a strong dependency on the intrinsic topological characteristics of the input stream, as quantified by the Periodicity Index ($PI$) in Fig. \ref{fig:PDs}. By comparing the performance gains in Table \ref{tab:results} with these $PI$ values, a clear Topological Resonance effect emerges. Specifically, the Mackey-Glass dataset, which possesses the highest $PI$ (0.1696) and exhibits prominent one-dimensional loops in its embedding space, achieves the most dramatic improvement. In the context of reservoir computing, these $H_1$ loops represent recurrent structures essential for capturing temporal dependencies. Our method explicitly strengthens these minimal representative cycles, allowing the reservoir's optimized digraph to structurally ``resonate'' with the data's cyclic attractor. This alignment facilitates more accurate detection and storage of periodic patterns, maximizing the model's forecasting accuracy.

However, our study further reveals that the method's effectiveness is not solely restricted to high-$PI$ datasets. A crucial observation is the dual-pathway mechanism of the GLMY theory-based optimization. While datasets like MSO, Lorenz, and NARMA (which have lower $PI$ values (0.0183–0.0269)) show more constrained improvements, the ETT dataset presents a compelling case. Despite having the lowest $PI$ (0.0135), ETT achieves a substantial performance gain. This phenomenon can be comprehensively understood by incorporating the Memory Capacity (MC) analysis from Table \ref{tab:mc}. Even when the ``Topological Resonance'' is weak due to indistinct loop features, the GLMY theory-based modification significantly expands the reservoir's fading memory (increasing ETT's MC from 18.06 to 51.04). Hence, the effectiveness of our method is the result of a joint influence: (1) Explicit Structural Alignment: directly matching 1-dimensional topological features for periodic data; (2) Functional Memory Extension: universally enhancing the network's information retention capacity via ring-structured optimization. Only when the method can effectively identify and utilize the topological foundation of the dataset, or significantly boost the requisite memory for the task, can it achieve significant improvements. This provides a generalized framework for understanding how structural optimization interacts with the complex dynamics of time-series data, offering valuable insights for integrating topological analysis with complex system modeling.

\subsubsection{Limitations}
There are some limitations of our method. First, in this study, we exclude consideration of bidirectional edges (referred to as bigons in the context of GLMY homology), and also, our approach neither captures nor modifies boundary triangles and boundary squares, as they do not constitute generators of GLMY homology, despite being cycles. In dense reservoir digraphs, as the connection density is large, boundary triangles can be numerical, making the number of representative cycles small. But for a sparse reservoir digraph, the boundary cycles are significantly less than the representative cycles; the majority of cycles can still be transformed into rings, thereby enhancing the reservoir's prediction capability. Hence, our method is more suitable for optimizing sparse reservoirs. And since many rings can share common edges (which means their corresponding fundamental cycle submatrices will share common columns), the increased number of rings is not proportional to improved orthogonality. It is still worthwhile studying more efficient and reasonable methods for improving orthogonality.

Additionally, providing a rigorous theoretical explanation for our method's efficacy on large-scale random sparse reservoirs remains challenging. This absence of strict theoretical bounds introduces the potential for performance variance. Given the intricate dynamics of reservoirs, certain pathological initializations may exist where modifying representative cycles inadvertently disrupts advantageous hidden sub-structures, resulting in marginal gains or localized performance degradation. In our future work, we will continue to study relevant theoretical explanations.
Moreover, the computational efficiency also becomes a concern for dense reservoirs or large reservoirs, as our method computes GLMY homology and modifies edge directions iteratively, which can be time-intensive. Our future research directions will focus on investigating the theoretical relationship between reservoir topology and computational capability, and developing accelerated algorithms for GLMY homology computation and our algorithm.

\section{Conclusions}\label{sec:conclusion}
In this study, we proposed a novel approach to optimize the structure of the reservoirs using GLMY homology theory. We begin by computing the one-dimensional GLMY homology of the reservoir digraph to obtain the corresponding minimal representative cycles. These cycles are then utilized to strategically modify the digraph structure in a compatible way, specifically increasing the number of rings within the reservoir. Experimental results demonstrate that this topological optimization leads to improved predictive performance across our tested datasets, and the proposed method performs better on datasets with significant 1-dimensional topological features.

\section*{CRediT authorship contribution statement}
\textbf{Yu Chen:} Conceptualization, Investigation, Formal analysis, Methodology, Software, Validation, Visualization, Writing – original draft, Writing – review \& editing.
\textbf{Shengwei Wang:} Methodology, Investigation, Software, Validation, Writing – original draft, Writing – review \& editing.
\textbf{Hongwei Lin:} Conceptualization, Methodology, Supervision, Funding acquisition, Writing – original draft, Writing – review \& editing.

\section*{Declaration of competing interest}
The authors declare that they have no known competing financial interests or personal relationships that could have appeared to influence the work reported in this paper.

\section*{Data availability}
The raw Electricity Transformer Dataset (ETT) is available at https://github.com/zhouhaoyi/ETDataset/tree/main.

\section*{Acknowledgments}
This work was partially supported by National Natural Science Foundation of China under Grant No. 62272406 and Leading Goose R\&D Program of Zhejiang under Grant No. 2024C01103.

\bibliographystyle{elsarticle-num}
\bibliography{reference.bib}

@article{jaeger2004harnessing,
  title={Harnessing nonlinearity: Predicting chaotic systems and saving energy in wireless communication},
  author={Jaeger, Herbert and Haas, Harald},
  journal={science},
  volume={304},
  number={5667},
  pages={78--80},
  year={2004},
  publisher={American Association for the Advancement of Science}
}

@article{jaeger2001echo,
  title={The “echo state” approach to analysing and training recurrent neural networks-with an erratum note},
  author={Jaeger, Herbert},
  journal={Bonn, Germany: German national research center for information technology gmd technical report},
  volume={148},
  number={34},
  pages={13},
  year={2001},
  publisher={Bonn}
}

@article{lukovsevivcius2009reservoir,
  title={Reservoir computing approaches to recurrent neural network training},
  author={Luko{\v{s}}evi{\v{c}}ius, Mantas and Jaeger, Herbert},
  journal={Computer science review},
  volume={3},
  number={3},
  pages={127--149},
  year={2009},
  publisher={Elsevier}
}

@article{rumelhart1986learning,
  title={Learning representations by back-propagating errors},
  author={Rumelhart, David E and Hinton, Geoffrey E and Williams, Ronald J},
  journal={nature},
  volume={323},
  number={6088},
  pages={533--536},
  year={1986},
  publisher={Nature Publishing Group UK London}
}

@article{venayagamoorthy2009effects,
  title={Effects of spectral radius and settling time in the performance of echo state networks},
  author={Venayagamoorthy, Ganesh K and Shishir, Bashyal},
  journal={Neural Networks},
  volume={22},
  number={7},
  pages={861--863},
  year={2009},
  publisher={Elsevier}
}

@article{grigor2012homologies,
  title={Homologies of path complexes and digraphs},
  author={Grigor'yan, Alexander and Lin, Yong and Muranov, Yuri and Yau, Shing-Tung},
  journal={arXiv preprint arXiv:1207.2834},
  year={2012}
}

@article{grigor2014homotopy,
	title={Homotopy theory for digraphs},
	author={Grigor'yan, Alexander and Lin, Yong and Muranov, Yuri and Yau, Shing-Tung},
	journal={arXiv preprint arXiv:1407.0234},
	year={2014}
}

@article{dey2022efficient,
  title={An efficient algorithm for 1-dimensional (persistent) path homology},
  author={Dey, Tamal K and Li, Tianqi and Wang, Yusu},
  journal={Discrete \& Computational Geometry},
  volume={68},
  number={4},
  pages={1102--1132},
  year={2022},
  publisher={Springer}
}

@article{edelsbrunner2008persistent,
  title={Persistent homology-a survey},
  author={Edelsbrunner, Herbert and Harer, John and others},
  journal={Contemporary mathematics},
  volume={453},
  number={26},
  pages={257--282},
  year={2008},
  publisher={Citeseer}
}

@book{edelsbrunner2010computational,
  title={Computational topology: an introduction},
  author={Edelsbrunner, Herbert and Harer, John},
  year={2010},
  publisher={American Mathematical Soc.}
}

@article{yao2019prediction,
  title={Prediction and identification of discrete-time dynamic nonlinear systems based on adaptive echo state network},
  author={Yao, Xianshuang and Wang, Zhanshan and Zhang, Huaguang},
  journal={Neural Networks},
  volume={113},
  pages={11--19},
  year={2019},
  publisher={Elsevier}
}

@article{skowronski2007noise,
  title={Noise-robust automatic speech recognition using a predictive echo state network},
  author={Skowronski, Mark D and Harris, John G},
  journal={IEEE Transactions on Audio, Speech, and Language Processing},
  volume={15},
  number={5},
  pages={1724--1730},
  year={2007},
  publisher={IEEE}
}

@article{shougat2021hopf,
  title={A Hopf physical reservoir computer},
  author={Shougat, Md Raf E Ul and Li, XiaoFu and Mollik, Tushar and Perkins, Edmon},
  journal={Scientific Reports},
  volume={11},
  number={1},
  pages={19465},
  year={2021},
  publisher={Nature Publishing Group UK London}
}

@article{wang2021stock,
  title={Stock market index prediction based on reservoir computing models},
  author={Wang, Wei-Jia and Tang, Yong and Xiong, Jason and Zhang, Yi-Cheng},
  journal={Expert Systems with Applications},
  volume={178},
  pages={115022},
  year={2021},
  publisher={Elsevier}
}

@article{cui2012architecture,
  title={The architecture of dynamic reservoir in the echo state network},
  author={Cui, Hongyan and Liu, Xiang and Li, Lixiang},
  journal={Chaos: An Interdisciplinary Journal of Nonlinear Science},
  volume={22},
  number={3},
  year={2012},
  publisher={AIP Publishing}
}

@inproceedings{gallicchio2019reservoir,
  title={Reservoir topology in deep echo state networks},
  author={Gallicchio, Claudio and Micheli, Alessio},
  booktitle={Artificial Neural Networks and Machine Learning--ICANN 2019: Workshop and Special Sessions: 28th International Conference on Artificial Neural Networks, Munich, Germany, September 17--19, 2019, Proceedings 28},
  pages={62--75},
  year={2019},
  organization={Springer}
}

@article{roeschies2010structure,
  title={Structure optimization of reservoir networks},
  author={Roeschies, Benjamin and Igel, Christian},
  journal={Logic Journal of IGPL},
  volume={18},
  number={5},
  pages={635--669},
  year={2010},
  publisher={Oxford University Press}
}

@inproceedings{chowdhury2018persistent,
  title={Persistent path homology of directed networks},
  author={Chowdhury, Samir and M{\'e}moli, Facundo},
  booktitle={Proceedings of the Twenty-Ninth Annual ACM-SIAM Symposium on Discrete Algorithms},
  pages={1152--1169},
  year={2018},
  organization={SIAM}
}

@inproceedings{chowdhury2019path,
  title={Path homologies of deep feedforward networks},
  author={Chowdhury, Samir and Gebhart, Thomas and Huntsman, Steve and Yutin, Matvey},
  booktitle={2019 18th IEEE International Conference On Machine Learning And Applications (ICMLA)},
  pages={1077--1082},
  year={2019},
  organization={IEEE}
}

@article{grigor2019homology,
  title={Homology of path complexes and hypergraphs},
  author={Grigor'yan, Alexander and Jimenez, Rolando and Muranov, Yuri and Yau, Shing-Tung},
  journal={Topology and its Applications},
  volume={267},
  pages={106877},
  year={2019},
  publisher={Elsevier}
}

@article{feng2024hypernetwork,
  title={Hypernetwork modeling and topology of high-order interactions for complex systems},
  author={Feng, Li and Gong, Huiying and Zhang, Shen and Liu, Xiang and Wang, Yu and Che, Jincan and Dong, Ang and Griffin, Christopher H and Gragnoli, Claudia and Wu, Jie and others},
  journal={Proceedings of the National Academy of Sciences},
  volume={121},
  number={40},
  pages={e2412220121},
  year={2024},
  publisher={National Academy of Sciences}
}

@inproceedings{chowdhury2020path,
  title={Path homology and temporal networks},
  author={Chowdhury, Samir and Huntsman, Steve and Yutin, Matvey},
  booktitle={International Conference on Complex Networks and Their Applications},
  pages={639--650},
  year={2020},
  organization={Springer}
}

@article{liu2023neighborhood,
  title={Neighborhood path complex for the quantitative analysis of the structure and stability of carboranes},
  author={Liu, Jian and Chen, Dong and Pan, Feng and Wu, Jie},
  journal={Journal of Computational Biophysics and Chemistry},
  volume={22},
  number={04},
  pages={503--511},
  year={2023},
  publisher={World Scientific}
}

@article{sun2012modeling,
  title={Modeling deterministic echo state network with loop reservoir},
  author={Sun, Xiao-chuan and Cui, Hong-yan and Liu, Ren-ping and Chen, Jian-ya and Liu, Yun-jie},
  journal={Journal of Zhejiang University SCIENCE C},
  volume={13},
  number={9},
  pages={689--701},
  year={2012},
  publisher={Springer}
}

@inproceedings{stockdill2016restricted,
  title={Restricted echo state networks},
  author={Stockdill, Aaron and Neshatian, Kourosh},
  booktitle={AI 2016: Advances in Artificial Intelligence: 29th Australasian Joint Conference, Hobart, TAS, Australia, December 5-8, 2016, Proceedings 29},
  pages={555--560},
  year={2016},
  organization={Springer}
}

@article{barabasi1999emergence,
  title={Emergence of scaling in random networks},
  author={Barab{\'a}si, Albert-L{\'a}szl{\'o} and Albert, R{\'e}ka},
  journal={science},
  volume={286},
  number={5439},
  pages={509--512},
  year={1999},
  publisher={American Association for the Advancement of Science}
}

@article{watts1998collective,
  title={Collective dynamics of ‘small-world’networks},
  author={Watts, Duncan J and Strogatz, Steven H},
  journal={nature},
  volume={393},
  number={6684},
  pages={440--442},
  year={1998},
  publisher={Nature Publishing Group}
}

@inproceedings{takens2006detecting,
  title={Detecting strange attractors in turbulence},
  author={Takens, Floris},
  booktitle={Dynamical Systems and Turbulence, Warwick 1980: proceedings of a symposium held at the University of Warwick 1979/80},
  pages={366--381},
  year={2006},
  organization={Springer}
}

@article{lin2021discrete,
	title={Discrete Morse theory on digraphs},
	author={Lin, Yong and Wang, Chong and Yau, Shing-Tung},
	journal={arXiv preprint arXiv:2102.10518},
	year={2021}
}

@inproceedings{grigor2018path,
	title={Path homology theory of multigraphs and quivers},
	author={Grigor’yan, Alexander and Muranov, Yuri and Vershinin, Vladimir and Yau, Shing-Tung},
	booktitle={Forum mathematicum},
	volume={30},
	number={5},
	pages={1319--1337},
	year={2018},
	organization={De Gruyter}
}

@article{jaeger2001short,
  title={Short term memory in echo state networks},
  author={Jaeger, Herbert},
  year={2001},
  publisher={GMD Forschungszentrum Informationstechnik}
}

@article{farkavs2016computational,
  title={Computational analysis of memory capacity in echo state networks},
  author={Farka{\v{s}}, Igor and Bos{\'a}k, Radom{\'\i}r and Gergel', Peter},
  journal={Neural Networks},
  volume={83},
  pages={109--120},
  year={2016},
  publisher={Elsevier}
}

@inproceedings{baranvcok2014memory,
  title={Memory capacity of input-driven echo state networks at the edge of chaos},
  author={Baran{\v{c}}ok, Peter and Farka{\v{s}}, Igor},
  booktitle={Artificial Neural Networks and Machine Learning--ICANN 2014: 24th International Conference on Artificial Neural Networks, Hamburg, Germany, September 15-19, 2014. Proceedings 24},
  pages={41--48},
  year={2014},
  organization={Springer}
}

@article{strauss2012design,
  title={Design strategies for weight matrices of echo state networks},
  author={Strauss, Tobias and Wustlich, Welf and Labahn, Roger},
  journal={Neural computation},
  volume={24},
  number={12},
  pages={3246--3276},
  year={2012},
  publisher={MIT Press}
}

@article{fasy2014confidence,
  title={Confidence sets for persistence diagrams},
  author={Fasy, Brittany Terese and Lecci, Fabrizio and Rinaldo, Alessandro and Wasserman, Larry and Balakrishnan, Sivaraman and Singh, Aarti},
  journal={The Annals of Statistics},
  volume={42},
  number={6},
  pages={2301},
  year={2014},
  publisher={Institute of Mathematical Statistics}
}

@inproceedings{deng2006complex,
  title={Complex systems modeling using scale-free highly-clustered echo state network},
  author={Deng, Zhidong and Zhang, Yi},
  booktitle={The 2006 IEEE international joint conference on neural network proceedings},
  pages={3128--3135},
  year={2006},
  organization={IEEE}
}

@article{kawai2019small,
  title={A small-world topology enhances the echo state property and signal propagation in reservoir computing},
  author={Kawai, Yuji and Park, Jihoon and Asada, Minoru},
  journal={Neural Networks},
  volume={112},
  pages={15--23},
  year={2019},
  publisher={Elsevier}
}

@article{montavon2018methods,
  title={Methods for interpreting and understanding deep neural networks},
  author={Montavon, Gr{\'e}goire and Samek, Wojciech and M{\"u}ller, Klaus-Robert},
  journal={Digital signal processing},
  volume={73},
  pages={1--15},
  year={2018},
  publisher={Elsevier}
}

@article{pathak2018model,
  title={Model-free prediction of large spatiotemporally chaotic systems from data: A reservoir computing approach},
  author={Pathak, Jaideep and Hunt, Brian and Girvan, Michelle and Lu, Zhixin and Ott, Edward},
  journal={Physical review letters},
  volume={120},
  number={2},
  pages={024102},
  year={2018},
  publisher={APS}
}

@article{lecun2015deep,
  title={Deep learning},
  author={LeCun, Yann and Bengio, Yoshua and Hinton, Geoffrey},
  journal={nature},
  volume={521},
  number={7553},
  pages={436--444},
  year={2015},
  publisher={Nature Publishing Group UK London}
}

@book{goodfellow2016deep,
  title={Deep learning},
  author={Goodfellow, Ian and Bengio, Yoshua and Courville, Aaron and Bengio, Yoshua},
  volume={1},
  number={2},
  year={2016},
  publisher={MIT press Cambridge}
}

@article{hornik1989multilayer,
  title={Multilayer feedforward networks are universal approximators},
  author={Hornik, Kurt and Stinchcombe, Maxwell and White, Halbert},
  journal={Neural networks},
  volume={2},
  number={5},
  pages={359--366},
  year={1989},
  publisher={Elsevier}
}

@article{gaier2019weight,
  title={Weight agnostic neural networks},
  author={Gaier, Adam and Ha, David},
  journal={Advances in neural information processing systems},
  volume={32},
  year={2019}
}

@article{zhang2024comprehensive,
  title={A comprehensive review of deep neural network interpretation using topological data analysis},
  author={Zhang, Ben and He, Zitong and Lin, Hongwei},
  journal={Neurocomputing},
  volume={609},
  pages={128513},
  year={2024},
  publisher={Elsevier}
}

@inproceedings{rieck2019neural,
  title={Neural Persistence: A Complexity Measure for Deep Neural Networks Using Algebraic Topology},
  author={Rieck, Bastian Alexander and Togninalli, Matteo and Bock, Christian and Moor, Michael and Horn, Max and Gumbsch, Thomas and Borgwardt, Karsten},
  booktitle={International Conference on Learning Representations (ICLR) 2019},
  year={2019}
}

@article{naitzat2020topology,
  title={Topology of deep neural networks},
  author={Naitzat, Gregory and Zhitnikov, Andrey and Lim, Lek-Heng},
  journal={Journal of Machine Learning Research},
  volume={21},
  number={184},
  pages={1--40},
  year={2020}
}

@incollection{lukovsevivcius2012practical,
  title={A practical guide to applying echo state networks},
  author={Luko{\v{s}}evi{\v{c}}ius, Mantas},
  booktitle={Neural Networks: Tricks of the Trade: Second Edition},
  pages={659--686},
  year={2012},
  publisher={Springer}
}

@article{ahmed2025experimental,
  title={An experimental evaluation, performance analysis, and improvement of water desalination system using optimized machine learning},
  author={Ahmed, Faizan and Mohammed, Nayeemuddin and Mewada, Hiren and Aziz, Mohd Sharizal Abdul and Khor, CY},
  journal={Process Integration and Optimization for Sustainability},
  pages={1--16},
  year={2025},
  publisher={Springer}
}

@article{ahmed2024experimental,
  title={Experimental and numerical investigation of an innovative desalination unit under laminar, transient, and turbulent flow conditions},
  author={Ahmed, Faizan and Aziz, Mohd Sharizal Abdul and Shaik, Feroz and Khor, CY},
  journal={Chemical Engineering Research and Design},
  volume={208},
  pages={683--694},
  year={2024},
  publisher={Elsevier}
}

@article{rodan2010minimum,
  title={Minimum complexity echo state network},
  author={Rodan, Ali and Tino, Peter},
  journal={IEEE transactions on neural networks},
  volume={22},
  number={1},
  pages={131--144},
  year={2010},
  publisher={IEEE}
}

@article{benedetto2003finite,
  title={Finite normalized tight frames},
  author={Benedetto, John J and Fickus, Matthew},
  journal={Advances in Computational Mathematics},
  volume={18},
  number={2},
  pages={357--385},
  year={2003},
  publisher={Springer}
}

@article{perea2015sliding,
  title={Sliding windows and persistence: An application of topological methods to signal analysis},
  author={Perea, Jose A and Harer, John},
  journal={Foundations of computational mathematics},
  volume={15},
  number={3},
  pages={799--838},
  year={2015},
  publisher={Springer}
}

@article{donoho2001uncertainty,
  title={Uncertainty principles and ideal atomic decomposition},
  author={Donoho, David L and Huo, Xiaoming and others},
  journal={IEEE transactions on information theory},
  volume={47},
  number={7},
  pages={2845--2862},
  year={2001}
}

@article{chen2025analyzing,
  title={Analyzing singular patterns in discrete planar vector fields via persistent path homology},
  author={Chen, Yu and Lin, Hongwei},
  journal={Computers \& Graphics},
  pages={104354},
  year={2025},
  publisher={Elsevier}
}

@inproceedings{haoyietal-informer-2021,
  author    = {Haoyi Zhou and
               Shanghang Zhang and
               Jieqi Peng and
               Shuai Zhang and
               Jianxin Li and
               Hui Xiong and
               Wancai Zhang},
  title     = {Informer: Beyond Efficient Transformer for Long Sequence Time-Series Forecasting},
  booktitle = {The Thirty-Fifth {AAAI} Conference on Artificial Intelligence, {AAAI} 2021, Virtual Conference},
  volume    = {35},
  number    = {12},
  pages     = {11106--11115},
  publisher = {{AAAI} Press},
  year      = {2021},
}

\end{document}